\pdfoutput=1

\documentclass[11pt]{article}
\usepackage[preprint]{coling}

\usepackage{times}
\usepackage{latexsym}

\usepackage[T1]{fontenc}

\usepackage[utf8]{inputenc}

\usepackage{microtype}

\usepackage{inconsolata}
\usepackage{graphicx}
\usepackage{booktabs}
\usepackage{multirow}
\usepackage{hyperref}
\usepackage{amsmath}
\usepackage{tcolorbox}
\usepackage{makecell}
\usepackage{pifont}
\usepackage{subfigure}
\definecolor{darkgreen}{RGB}{0,100,0}
\title{\textsc{BanTH}: A Multi-label Hate Speech Detection \\Dataset for Transliterated Bangla}

\author{Fabiha Haider$^{\text{\textbf{1}}}$\thanks{Equal Contribution}, Fariha Tanjim Shifat$^\text{\textbf{1}}$\footnotemark[1], Md Farhan Ishmam$^{\text{\textbf{1,2}}}$\footnotemark[1], \textbf{Deeparghya}\\
\textbf{Dutta Barua}$^\text{\textbf{1}}$,\textbf{~Md Sakib Ul Rahman Sourove}$^\text{\textbf{1}}$\textbf{,~Md Fahim}$^\text{\textbf{1,3}}$\textbf{,~Md Farhad Alam}$^\text{\textbf{1}}$ \\
  $^\text{\textbf{1}}$Research and Development, Penta Global Limited, Bangladesh\\
    $^\text{\textbf{2}}$Islamic University of Technology, Bangladesh \\
  $^\text{\textbf{3}}$CCDS Lab, Independent University, Bangladesh \\
  \texttt{pdcsedu@gmail.com}, 
  \texttt{fahimcse381@gmail.com}
}

\begin{document}
\maketitle

\begin{abstract}
The proliferation of transliterated texts in digital spaces has emphasized the need for detecting and classifying hate speech in languages beyond English, particularly in low-resource languages. As online discourse can perpetuate discrimination based on target groups, e.g. gender, religion, and origin, multi-label classification of hateful content can help in understanding hate motivation and enhance content moderation. 
While previous efforts have focused on monolingual or binary hate classification tasks, no work has yet addressed the challenge of multi-label hate speech classification in transliterated Bangla. We introduce \textsc{BanTH}, the first multi-label transliterated Bangla hate speech dataset. The samples are sourced from YouTube comments, where each instance is labeled with one or more target groups, reflecting the regional demographic. We propose a novel translation-based LLM prompting strategy that translates or transliterates under-resourced text to higher-resourced text before classifying the hate group(s). Experiments reveal further pre-trained encoders achieving state-of-the-art performance on the \textsc{BanTH} dataset while translation-based prompting outperforms other strategies in the zero-shot setting. We address a critical gap in Bangla hate speech and set the stage for further exploration into code-mixed and multi-label classification in underrepresented languages.
\end{abstract}

\noindent\textcolor{red}{\textbf{Content Warning:} This article contains examples of hateful content.}\\

\noindent\textbf{Note: }Throughout the work, we use the term \emph{Bangla} referring to both \emph{Bangla} and the endonym \emph{Bengali}. The terms denote the same language, primarily spoken by people of the West Bengal region of India and the vast majority of Bangladesh. 

\section{Introduction}


\begin{figure}
    \centering
    \includegraphics[width=0.45\textwidth]{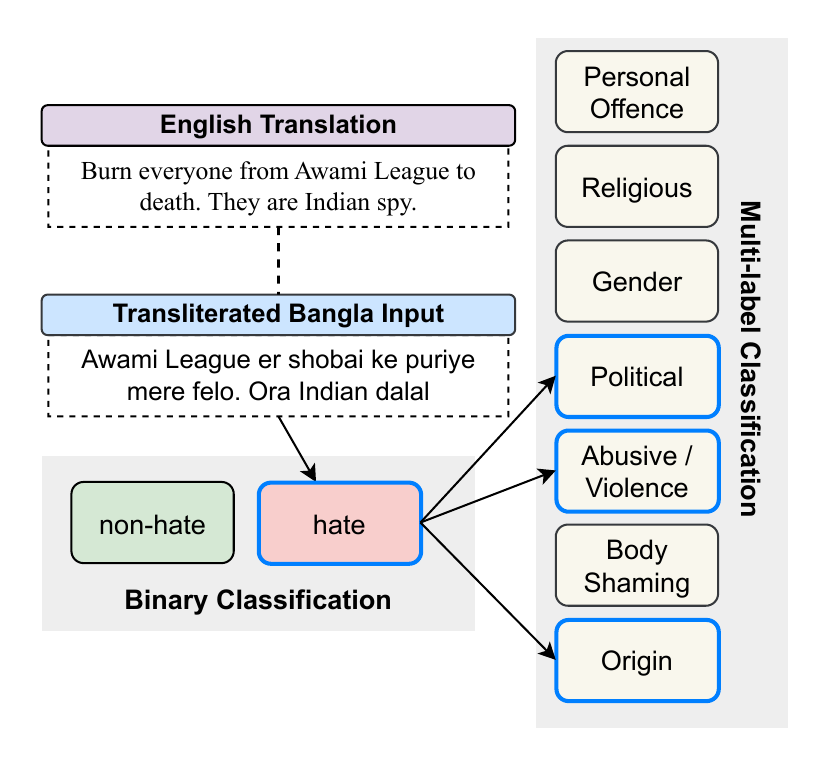}
    \caption{Binary and Multi-label Classification on an example Transliterated Bangla sentence from the \textsc{BanTH} dataset along with its corresponding English translation.}
    \label{fig:example}
\end{figure}

\begin{figure*}[ht]
    \centering
    \includegraphics[width=\textwidth]{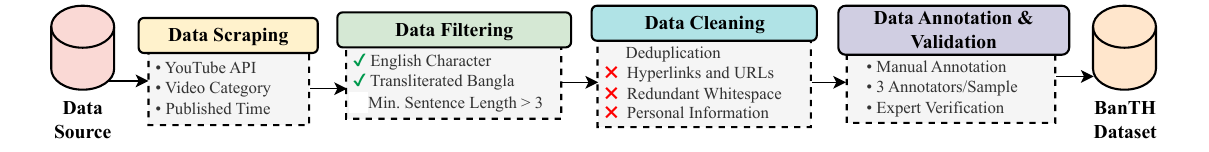}
    \caption{Overview of the \textsc{BanTH} dataset creation pipeline.}
    \label{fig:DatasetCreation}
\end{figure*}








The ever-expanding digital landscape that promises to improve social cohesion has been a breeding ground for hate speech \cite{castano2021internet}. Hate speech is defined as any form of language that targets, attacks, or incites implicit or explicit forms of hatred or violence against groups, based on specific characteristics, e.g. physical appearance, religion, ethnic origin, and gender identity \cite{papcunova2023hate}. Hate speech can potentially inflict societal harm by promoting division among communities, exacerbating mental health problems, and inciting violence \cite{sahoo2024indicconan}. The lack of hate speech moderation cultivates an environment of intolerance and magnifies the negative impact on the target communities \cite{hangartner2021empathy}. Categorizing hate speech requires a multi-faceted approach to capture overlapping hate categories and gain a granular understanding of the underlying motives behind the hateful discourse.

A prevalent informal form of online communication for non-English languages uses transliterated texts, i.e. writing a language in foreign script. Transliteration is predominant in Bangla, where Latin characters are used to produce a colloquial form of Bangla texts that does not strictly adhere to the original linguistic rules. Despite boasting over a quarter of a billion speakers worldwide, Bangla remains a low-resource language in terms of Natural Language Processing (NLP) tasks \cite{mahfuz-etal-2025-late}. The challenges are compounded when working with the even more under-resourced transliterated Bangla, which dominates online spaces due to the widespread familiarity of users with English keyboard layouts \cite{fahim-etal-2024-banglatlit}.

Although transliteration facilitates easier typing and cross-linguistic communication, it also complicates the task of automated hate speech detection due to inconsistent spelling and structure, absence of grammar, mixing with English, and the loss of script-specific features \cite{9290630}. While standard practices have employed transformer-based encoders \cite{devlin-etal-2019-bert} for automated hate speech detection, the recent surge in popularity of Large Language Models (LLMs) has positioned them as a viable option in hate speech NLP, particularly in the zero-shot setting. While most of the advancements have been centered on English and other high-resource languages, transliterated Bangla has limited research on hate or hate-like speech detection \cite{jahan2019abusive,raihan2023offensive} and LLM-based methods \cite{shibli2023automatic}, with no work addressing the intersection of these areas. 

Addressing the aforementioned research gap, our contribution can be summarized as follows:
\begin{itemize}
    \item We propose \textsc{BanTH}, the first multi-label hate speech detection dataset on transliterated Bangla with 37,350 samples.
    \item We establish several encoder-based baselines, with encoders further pre-trained on transliterated Bangla achieving state-of-the-art accuracy on our dataset.
    \item We explore zero-shot and few-shot prompting techniques using state-of-the-art Large Language Models and introduce a novel translation-based prompting strategy that outperforms existing methods on our dataset.
\end{itemize}

\begin{table*}[ht]
\centering
\resizebox{0.85\textwidth}{!}{
\begin{tabular}{ l | c c c | c c c |c}
\toprule
\textbf{Hate Category} &  \makecell[c]{ \#\textbf{News}\\ \textbf{\& Politics} } & \makecell[c]{ \#\textbf{People}\\ \textbf{\& Blogs} } & \makecell[c]{\#\textbf{Enter-}\\\textbf{tainment} }
& \textbf{\#Train} & \textbf{\#Val} & \textbf{\#Test} & \textbf{\#Total} \\ 
\midrule
Political &	1005 & 174 & 21 & 958 & 120 & 122 & 1200\\
Religious &	109 & 30 & 0 & 111 & 14 & 14 & 139\\
Gender & 151 & 38 & 4 & 154 & 19 & 20 & 193\\
Personal Offense & 5883 & 784 & 297 & 5572 & 696 & 696 & 6964 \\
Abusive/Violence & 2491 & 276 & 121 & 2312 & 287 & 289 & 2888\\
Origin & 201 & 11 & 2 & 171 & 21 & 22 & 214\\
Body Shaming & 56 & 29 & 1& 68 & 8 & 10 & 86 \\
\bottomrule
\end{tabular}
}
\caption{Distribution of multi-label hate speech among the YouTube video categories and dataset splits of \textsc{BanTH}.}
\label{tab:video-categorywise-multilabel}
\end{table*}

\begin{table*}[h!]
    \centering
    \resizebox{0.985\textwidth}{!}{
    \begin{tabular}{l l c l}
         \toprule
         \textbf{Time Period} & \textbf{Event} & \textbf{\#Samples} & \textbf{Major Hate Categories} \\ \midrule

        Dec'19 - Dec'22	& COVID-19 & 	878	 & Abusive/Violence, Personal \\
Dec'22	& Woman dragged by car to death&	1245	& Personal, Abusive/Violence  \\
Feb'24	& BCL leader rapes woman in JU &	1542	& Personal, Abusive/Violence, Political \\
July'24 - Aug'24 &	Quota reform movement&	4522 &	Political, Abusive/Violence, Personal \\
Aug'24 &	Anti-Hindu Violence &	1081 & Religious, Abusive/Violence \\
         \bottomrule
    \end{tabular}
    }
    \caption{Events covered in \textsc{BanTH} with the number of hate samples and associated major hate categories.}
    \label{tab:eventWiseHate}
\end{table*}

\section{\textsc{BanTH} Dataset}

The \textsc{BanTH} dataset comprises 37350 samples, each initially classified into binary labels of \textit{Hate} or \textit{Non-Hate}. To effectively capture the target group, the hate samples are further multi-labeled as \textit{Political, Religious, Gender, Personal Offense, Abusive/Violence, Origin,} and \textit{Body Shaming}. The description of target groups is provided in \S\ref{app:target_descriptopn} and the dataset creation flowchart is given in Fig. \ref{fig:DatasetCreation}.

\subsection{Data Sourcing}
We construct the dataset by scraping user comments from public YouTube videos using the YouTube API\footnote{\url{ https://developers.google.com/youtube/v3/getting-started}}. The scraped comments include three categories of videos: ``News \& Politics'', ``People \& Blogs'', and ``Entertainment'', totaling 26 different YouTube channels, outlined in Tab. \ref{tab:dataset-stats}. The contents often cover a wider range of topics, including political analysis, interviews, talk shows, product reviews, and travel videos, regionally relevant to West Bengal of India and Bangladesh.

The scraped comments were collected from January 2020 to July 2024, a time frame that captures significant recent events, e.g. the COVID-19 pandemic and the Bangladesh quota reform movement\footnote{\url{https://en.wikipedia.org/wiki/2024_Bangladesh_quota_reform_movement}}. Such events led to increased social media activity. The variety of content appeals to a diverse audience with a wide range of interests and perspectives. The label-wise distribution of video categories is elaborated in Tab. \ref{tab:video-categorywise-multilabel} and event-wise major hate categories are described in Tab. \ref{tab:eventWiseHate}.








\subsection{Data Filtering}
We primarily filtered comments written in Roman scripts and discarded comments in other transliterated languages, \textit{e.g.} Hindi, during annotation. The scraped comments were further filtered using dictionary-based language identification to ensure only text from the desired language remains. Finally, texts with a length greater than or equal to three were kept in the dataset.

\subsection{Data Cleaning}
During data cleaning, we removed URLs using regular expressions as they were irrelevant to our task. We also removed comments that were verbatim duplicates of others. Personal information like addresses and contact numbers was also removed during annotation. 

\subsection{Data Annotation}
We hired four annotators and two domain experts for the data annotation process. All the annotators were high school graduates and native Bangla speakers. They had extensive exposure to social media content and actively used transliterated Bangla text. The annotation process was supervised by domain experts with experience in working with transliterated Bangla text, who provided guidance and ensured clear and proper annotation guidelines, outlined in Appendix \ref{sec:annotation-guidelines}. Both the annotators and domain experts received monetary compensation.

Each data sample was annotated by three annotators to ensure consistency and capture a shared perspective. The annotators were first instructed to classify a sample as hate or non-hate. The resulting hate sample was then multi-labeled according to the target group descriptions given in \S\ref{app:target_descriptopn}. In cases where annotators encountered difficulties in reaching a consensus or understanding the text, the domain experts resolved the issues.

\begin{table}[ht]
\centering
\resizebox{.4\textwidth}{!}{
\begin{tabular}{lcc}
\toprule
    \textbf{Fleiss' Kappa} & \makecell[c]{\textbf{Expert-}\\ \textbf{Annotator}} & \makecell[c]{\textbf{Inter-} \\\textbf{Annotator}}\\
    \hline
   Binary Labeling & 0.75 & 0.71\\
   Political & 0.62 & 0.72\\
   Religious & 0.68 & 0.66\\
   Gender & 0.64 & 0.68\\
   Personal Offense & 0.70 & 0.75\\
   Abusive/Violence & 0.72 & 0.71\\
   Origin &  0.68 & 0.72\\
   Body Shaming & 0.64 & 0.66 \\
   \hline
\end{tabular}}
\caption{Expert-annotator and inter-annotator Fleiss' Kappa Score across hate categories.}
\label{tab:kappaScore}
\end{table} 

\subsection{Data Validation}
Following the initial phase of data annotation, the annotators verified whether any samples remained unlabeled, samples were mislabeled, or if there were instances where a sample was labeled as hate without the corresponding multi-label, and vice versa. Additionally, they ensured that all samples were in transliterated Bangla only, as samples in other languages are beyond the scope of our work. The agreement scores, measured using the Fleiss' Kappa, are shown the Table \ref{tab:kappaScore}.  We found an inter-annotator agreement score of 0.53, indicating a moderate agreement across the hate categories \cite{islam-etal-2021-sentnob-dataset}.


\begin{table}[t]
\centering
\resizebox{.35\textwidth}{!}{
\begin{tabular}{ l r r}
\toprule
\textbf{Source Category} &  \#\textbf{Hate} & \#\textbf{Non-hate} \\ 
\midrule
News \& Politics & 8745 & 23695 \\ 
People \& Blogs & 1125 & 2889 \\ 
Entertainment & 412 & 584\\ 
\midrule
{\textbf{Total}} & 10282  & 27167 \\ 
\midrule
\multicolumn{2}{l}{\textbf{Dataset Statistics}} \\ 
\midrule
\multicolumn{2}{l}{Min Character Count} &7\\
\multicolumn{2}{l}{Max Character Count} &1951 \\
Average Length & &58.09\\
\multicolumn{2}{l}{Standard Deviation} & 65.13 \\
Total Words & &379411 \\
Average Words & &10.16 \\
Min Word Count & &3\\
Max Word Count & &368\\
Unique Words & &75977\\
\midrule
\textbf{\textsc{BanTH}  split}   \\ 
\midrule
- Train &  & 29879  \\
- Val &   & 3736  \\
- Test & & 3735   \\ 
\midrule
{\textbf{Total}} &  & 37350  \\ 
\bottomrule
\end{tabular}
}
\caption{Dataset statistics of our \textsc{BanTH} dataset.}
\label{tab:dataset-stats}
\end{table}

\subsection{Dataset Statistics}
\begin{figure}[t]
    \centering
    \includegraphics[width=.475\textwidth]{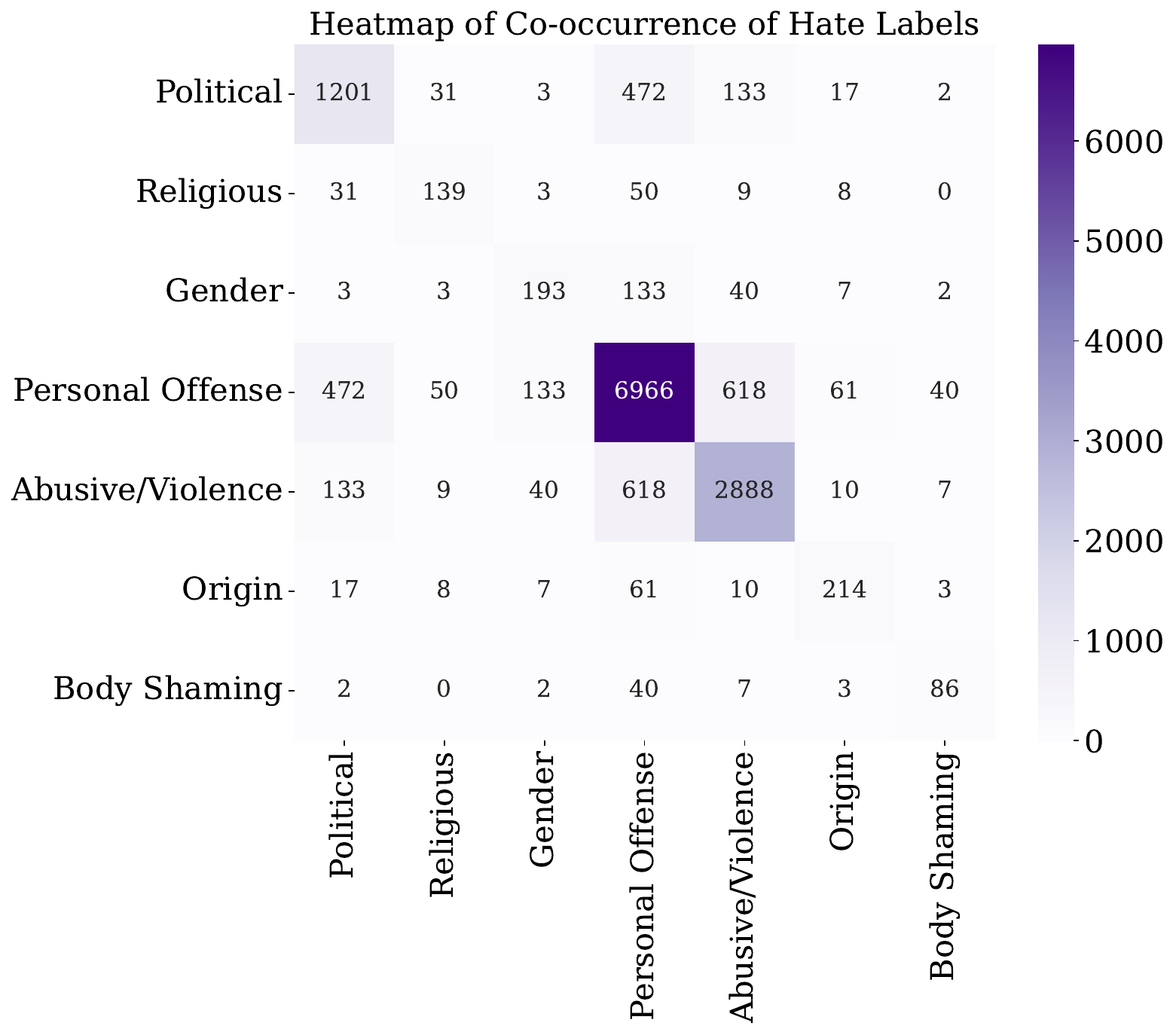}
    \caption{Co-occurrence of multi-label hate \textsc{BanTH}.}
    \label{fig:labeling-heatmap}
\end{figure}

The train, test, and val dataset splits maintain the standard 80:10:10 ratio. While splitting, stratification was followed to maintain a uniform distribution of hate and non-hate labels in multi-label categories. Tab. \ref{tab:video-categorywise-multilabel} shows the distribution across video categories and dataset splits while general statistics is given Tab. \ref{tab:dataset-stats}. Fig. \ref{fig:labeling-heatmap} visualizes the multi-label co-occurrence within the \textsc{BanTH} dataset, illustrating the frequency of label pairs appearing together in the text. Hate categories like \textit{Personal} and \textit{Political} frequently co-occur with other categories while the opposite is true for categories like \textit{Religious} and \textit{Body Shaming}.

\section{Methodology}
The experiments are carried out in two stages: binary classification of hate speech and multi-label (target group) classification of the predicted hate speech. Our baselines can be divided into two approaches: fine-tuning LMs on our dataset and prompting GPT-based LLMs using different prompting and shot settings.

\begin{table*}[htpb]
\centering
\resizebox{.9\textwidth}{!}{
\begin{tabular}{ll}
\toprule
\textbf{Prompt-Strategy} &\textbf{Prompt} \\
\midrule
Non-Explanatory & Base-prompt \\
Chain of Though (CoT) & Base-prompt + "Let's think step by step" \\
 Explanation-based (Exp) & Base-prompt + "Explain why"  \\
 HARE & Base-prompt + "Let's explain step by step" \\
 Why [Positive]& Base-prompt + "Explain why the comment is positive" \\
 Why [Negative]& Base-prompt + "Explain why the comment is negative" \\
  Translation Based \texttt{[BAN]}& "Translate the following transliterated text into standard Bangla" + Prompt-Strategy\\
 Translation Based \texttt{[ENG]}& "Translate the following transliterated text into standard English" + Prompt-Strategy\\
\bottomrule
\end{tabular}
}
\caption{The prompting strategies used in our benchmarks. We consider binary/multi-label classification and zero-shot/few-shot approaches along with their corresponding prompts.}
\label{tab:prompt-variation}
\end{table*}

\subsection{Further Pretraining}
Pretrained LMs such as BERT \cite{devlin-etal-2019-bert} and BanglaBERT \cite{banglabert} are typically trained on general and formal texts. However, these models often lack transliterated samples in their pretraining datasets, essentially becoming a long-tail distribution problem. To enhance the performance, we employ \textit{Further Pretraining (FPT)} \cite{qiu2021different, gururangan2020don}, utilizing the Masked Language Modeling (MLM) loss \cite{devlin-etal-2019-bert, zhuang-etal-2021-robustly} as our pretraining objective.

Specifically, for a given sentence \( S \), we represent the input tokens as \( X = (x_1, x_2, \ldots, x_n) \). We randomly mask 15\% of these tokens with the \textit{[MASK]} token, resulting in a masked sequence denoted as \( \tilde{X} \). Let \( M \subseteq \{1, 2, \ldots, n\} \) represent the set of indices of the masked tokens. We then input \( \tilde{X} \) into the language model \( f_{LM} \). For each masked token \( x_i \) (where \( i \in M \)), the model generates predicted probabilities \( P(x_i | \tilde{X}) \). The loss for each masked token is calculated as  \(
L_i = -\log P(x_i | \tilde{X})
\). The overall loss across all masked tokens is given by:
\[
L_{\text{MLM}} = -\frac{1}{|M|} \sum_{i \in M} \log P(x_i | \tilde{X})
\]

We use the encoders further pre-trained on BanglaTLit-PT corpus\cite{fahim-etal-2024-banglatlit}, which contains 243k unlabeled Bangla transliterated texts. The encoders are referred to as TB-Encoders (Transliterated Bangla Encoders).

\subsection{LM Finetuning}
We input a sentence \( \mathbf{S} \) into a pre-trained language model \( f_{\text{LM}} \) to obtain layer-wise contextual representations \( \mathbf{H} =   \{\mathbf{h}_i^l\}_{l=1}^L \), where \( \mathbf{H} = f_{\text{LM}}(S)\) and
\( L \) = no. of layers in \( f_{\text{LM}} \). We use the representation from the last layer's [CLS] token, $h^L_{\text{CLS}}$, as the sentence representation, which is then passed through a Multi-Layer Perceptron (MLP) for classification. The transformation is defined as follows:
\[
z = W_2 \cdot (\text{ReLU}(W_1 \cdot h^L_{\text{CLS}} + b_1)) + b_2
\]

The resulting representation \( z \) is then employed for calculating the loss. The TB-Encoders are similarly fine-tuned.







\subsection{Prompting Strategy}
We adopt one of the following strategies: Why Positive/Why Negative \cite{wang2023evaluatinggpt3generatedexplanations}, HARE \cite{yang2023hareexplainablehatespeech}, non-explanatory, explanation-based, CoT \cite{wei2022chain} based or translation based. An overview of the prompts has been reported in Tab. \ref{tab:prompt-variation}. We consider GPT 4o \cite{achiam2023gpt} and GPT-3.5 \cite{brown2020language} in the prompting experiments and have designed separate base prompts for binary and multi-label classification. Each of these base prompts is then extended to one of the prompting strategies. We consider both zero-shot \cite{radford2019language} and few-shot \cite{brown2020language} prompting techniques, reported in App. \ref{sec: prompts}. The prompt variations are detailed below:\\

\noindent \textbf{Vanilla/Base.} The base prompting strategy where we only ask the LLM to do binary or multi-label classification. The prompt also includes the hate speech definition, the target groups, and their definition in multi-label classification, key indicators of hate, and labeling instructions.\\

\noindent \textbf{Explanation Based.} We include the phrase \emph{Explain why} with the base prompt and ask the LLM to show its reasoning.\\

\noindent \textbf{HARE}. We add \emph{Let's explain step by step}, as adapted in \citet{yang2023hareexplainablehatespeech}, along with the base prompt and ask the LLM to return its reasoning.\\

\noindent \textbf{Why Positive Why Negative.} Following \citet{wang2023evaluatinggpt3generatedexplanations}, we adopt the approach by adding the question \emph{why the given text is positive or negative} appending to the base prompt in this variation.\\

\noindent \textbf{CoT-Based.} Following \citet{wei2022chain}, we add \emph{Let's think step by step} along with the base prompt.\\

\begin{table}[b]
    \centering
    \resizebox{.4\textwidth}{!}{
    \begin{tabular}{lccc}
    \toprule
    \textbf{Model}&\textbf{Label}&\textbf{Macro-F1 $\uparrow$}&\textbf{Acc. $\uparrow$}\\ \midrule
    \multirow{2}{*}{GPT 3.5}& Positive & 62.48 & 69.03 \\
     & Negative & 62.71 & 66.77 \\
    \multirow{2}{*}{GPT 4o} & Positive & 63.33 & 74.91 \\
    & Negative & 69.98 & 75.29 \\ \bottomrule
    \end{tabular}}
    \caption{Performance of Why prompting on the BanTH test split for binary classification.}
    \label{tab:whyPrompting}
\end{table}

\begin{table*}[htpb]
    \centering
    \resizebox{.99\textwidth}{!}{
    \begin{tabular}{lccccc}
        \toprule
        \multirow{2}{*}{\textbf{Models}} & \multicolumn{2}{c}{\textbf{Binary}} & \multicolumn{3}{c}{\textbf{Multi-label}} \\
        \cmidrule(lr){2-3} \cmidrule(lr){4-6}
        & \textbf{Macro-F1 $\uparrow$} & \textbf{Acc. $\uparrow$} & \textbf{Macro-F1 $\uparrow$} & \textbf{Subset Acc. $\uparrow$} & \textbf{Hamming Loss $\downarrow$} \\
        \midrule
        \multicolumn{6}{c}{\textbf{Language Model (LM) Fine-tuning}}\\
        \midrule
    \textbf{Bangla LM}\\
     \quad BanglaBERT \cite{banglabert} & 76.50 & 81.04 & 20.82 & 54.08 & 7.26 \\
    \quad BanglishBERT \cite{banglabert} &  75.07 & 80.62 & 20.61 & 52.74 & 7.42 \\
    \quad BanglaHateBERT \cite{banglahatebert} & 70.92  & 77.54 & 11.34 & 49.83 & 7.97  \\ 
     \quad VAC-BERT \cite{vacbert} & 74.19  & 79.76 & 18.45 & 52.56 & 7.50 \\  
    \midrule\textbf{Indian LM}\\ 
    \quad MuRIL \cite{muril} & 75.29 & 80.83 & 14.58 & 53.98 & 7.55 \\ 
    \quad IndicBERT \cite{indicbert} & 74.51 & 80.48 & 13.39 & 51.56 & 7.88 \\
    \midrule\textbf{Multilingual LM}\\
    \quad mBERT \cite{devlin-etal-2019-bert} & 74.97 & 80.37 & 28.24 & 52.19 & 7.78 \\
    \quad XLM-R \cite{xlm-roberta}& 77.35 & 81.37 & 29.29 & 53.28 & 7.23\\
    \midrule\textbf{Character-based LM} \\
    \quad CharBERT \cite{charbert} & 76.61 & 80.91 & 19.66 & 53.21 & 7.44 \\ \midrule
   \multicolumn{6}{l}{\textbf{Transliterated Bangla (TB) LM }\cite{fahim-etal-2024-banglatlit}}\\
    \quad TB-BERT & 76.27 & 79.25 & \textbf{30.17} & 54.19 & \textbf{7.18} \\
    \quad TB-mBERT & \textbf{77.36} & \textbf{82.57} & 27.07 & \textbf{54.71} & 7.28 \\
    \quad TB-XLM-R & 77.04 & 81.29 & 29.04 & 52.86 & 7.26 \\
    \quad TB-BanglaBERT & 77.12 & 81.61 & 22.52 & 53.97 & 7.37 \\
    \quad TB-BanglishBERT & 77.12 & 81.39 & 21.41 & 52.79 & 7.42 \\ 
    \midrule
    
    \multicolumn{6}{c}{\textbf{Large Language Model (LLM) Prompting}} \\ \midrule
    \multicolumn{6}{l}{\textbf{Vanilla/Base Prompting}}\\
    \quad GPT 3.5 \cite{brown2020language} & 61.44  & 64.02 & 24.12  & 18.27  & 19.77 \\
    \quad GPT 4o \cite{achiam2023gpt} & 70.05  & 74.30 & 36.07  & 22.60  & 16.91\\
     \quad GPT 3.5 + Few-shot& 61.85  & 65.65 & 23.86 & 16.85 & 20.36  \\
   \quad GPT 4o + Few-shot & 68.77  & 74.18 & \textbf{39.53}  & \textbf{26.76} & \textbf{14.16}\\
    \midrule\multicolumn{6}{l}{\textbf{Chain of Thought (CoT) Prompting}}\\
    \quad GPT 3.5 + CoT& 61.87  & 65.77 & 25.61 & 20.34  & 19.28\\
     \quad GPT 4o + CoT& 69.87  & 74.14 & 35.97  & 22.60  & 16.64  \\
    \quad GPT 3.5 + CoT + Few-shot & 63.30  & 67.80 & 28.35 & 20.15  & 19.15\\
        \quad GPT 4o + CoT + Few-shot & 69.50  & 74.94 & 36.49  & 21.10 & 17.36\\
    \midrule\multicolumn{6}{l}{\textbf{Explanation-based (Exp) Prompting}}\\
    \quad GPT 3.5 + Exp & 61.69  & 65.16 & 22.61  & 17.33 & 19.84\\
    \quad GPT 4o + Exp & 69.91  & 74.16 & 32.58  & 19.12 & 17.94  \\
     \quad GPT 3.5 + Exp + Few-shot & 62.69  & 66.60 & 25.00 & 18.36  & 19.57 \\
     \quad GPT 4o + Exp + Few-shot & \textbf{70.70} & \textbf{75.15} &  35.61  & 21.33  & 17.15 \\
     \midrule\multicolumn{6}{l}{\textbf{HARE Prompting}}\\

    \quad GPT 3.5 + HARE & 62.64  & 66.77 & 25.99 & 20.90  & 18.94 \\
     \quad GPT 4o + HARE& 69.71  & 74.75 & 36.67  & 20.80  & 17.10  \\
    \quad GPT 3.5 + HARE + Few-shot &  62.20 & 66.72 &  24.68  & 20.06 & 19.18  \\
    \quad GPT 4o + HARE + Few-shot & 69.12  & 74.75 & 34.06  & 21.59 & 17.37\\

    \midrule

    \textbf{Translation Prompting (Ours)}\\
    \quad GPT 4o + Translation \texttt{[BAN]}  & 69.63  & 72.70 & 35.72 & 23.28 & 16.73\\
    \quad GPT 4o + Translation \texttt{[ENG]} & 69.38  & 72.96 & 34.28 & 20.72 & 16.91 \\
    \quad GPT 4o + Translation \texttt{[BAN]} + CoT  & 69.01  & 72.66 & 36.22  & 21.35 & 17.36\\
    \quad GPT 4o + Translation \texttt{[ENG]} + CoT  &  69.74 & 74.87 & 35.94  & 21.19 & 16.91 \\
    \quad GPT 4o + Translation \texttt{[BAN]} + Exp  & \underline{70.34}  & 73.65 & 36.20  & 21.54 & 16.46\\
    \quad GPT 4o + Translation \texttt{[ENG]} + Exp  & 70.19  & 73.84 &  \underline{36.25} & \underline{23.92}& \underline{16.23} \\
    
        \bottomrule
    \end{tabular}
    }
       \caption{Model Benchmarking on the test split of \textsc{BanTH} Dataset for Binary and Multi-label Classification. For LLMS, the results show the zero-shot and few-shot performance for each prompting strategy. \textbf{Bold} represents overall the best results whereas \underline{Underline} represents the best performing results in \textit{zero-shot} settings.}
    \label{tab:ModelBenchmarking}
\end{table*}

\noindent \textbf{Translation Based.} Our proposed prompting strategy requires LLMs to first convert Bengali transliterated text into standard Bangla or English before proceeding with the usual classification. 
We hypothesize translating or transliterating an under-resourced representation to a higher-resourced representation should help mitigate the problem of processing under-represented texts, thereby mitigating the long-tail distribution problem. Furthermore, LLMs are relatively better at simpler \texttt{seq2seq} tasks such as transliterating or translating.
We mix translation-based prompting with Non-Explanatory, Explanatory, and CoT-based prompting techniques.\\

\section{Experimental Result}
The experimental setup is reported in App. \ref{app:exp_setup}. We evaluate fine-tuning vanilla pre-trained and further pre-trained LMs and different prompting strategies on LLMs. The evaluation metric details have also been provided in App. \ref{app:performance_metrics}.

\subsection{Fine-Tuning LMs Baseline}
For the binary classification, TB-mBERT achieves the best performance across all metrics. In multi-label classification, TB-BERT outperforms the other methods based on macro-F1 and hamming loss but is outperformed by TB-mBERT based on subset accuracy. Overall, the TB encoders have better performance across all metrics in both binary and multi-label classification tasks.

\subsection{Prompt Based LLM Performance}
Among prompt-based LLMs, GPT-4o + Few-shot performed the best for multi-label classification, achieving the highest macro-F1 (39.53\%), subset accuracy (26.76\%), and lowest hamming loss (14.16\%). In binary classification, GPT 4o+Exp+Few-shot has the highest macro-F1 (70.70\%). In zero-shot settings, our translation prompting strategy achieves the best performance across all metrics by a good margin (\textasciitilde{8}\% improvement over vanilla prompting). Additionally, GPT-4 consistently outperforms GPT-3.5 across all experiments and prompting variations, yielding improvements of 8-10\%. We also find that generating explanations can occasionally boost performance, but may also lead to worse results.




\begin{figure}[t]
    \centering
    \includegraphics[width=.45\textwidth]{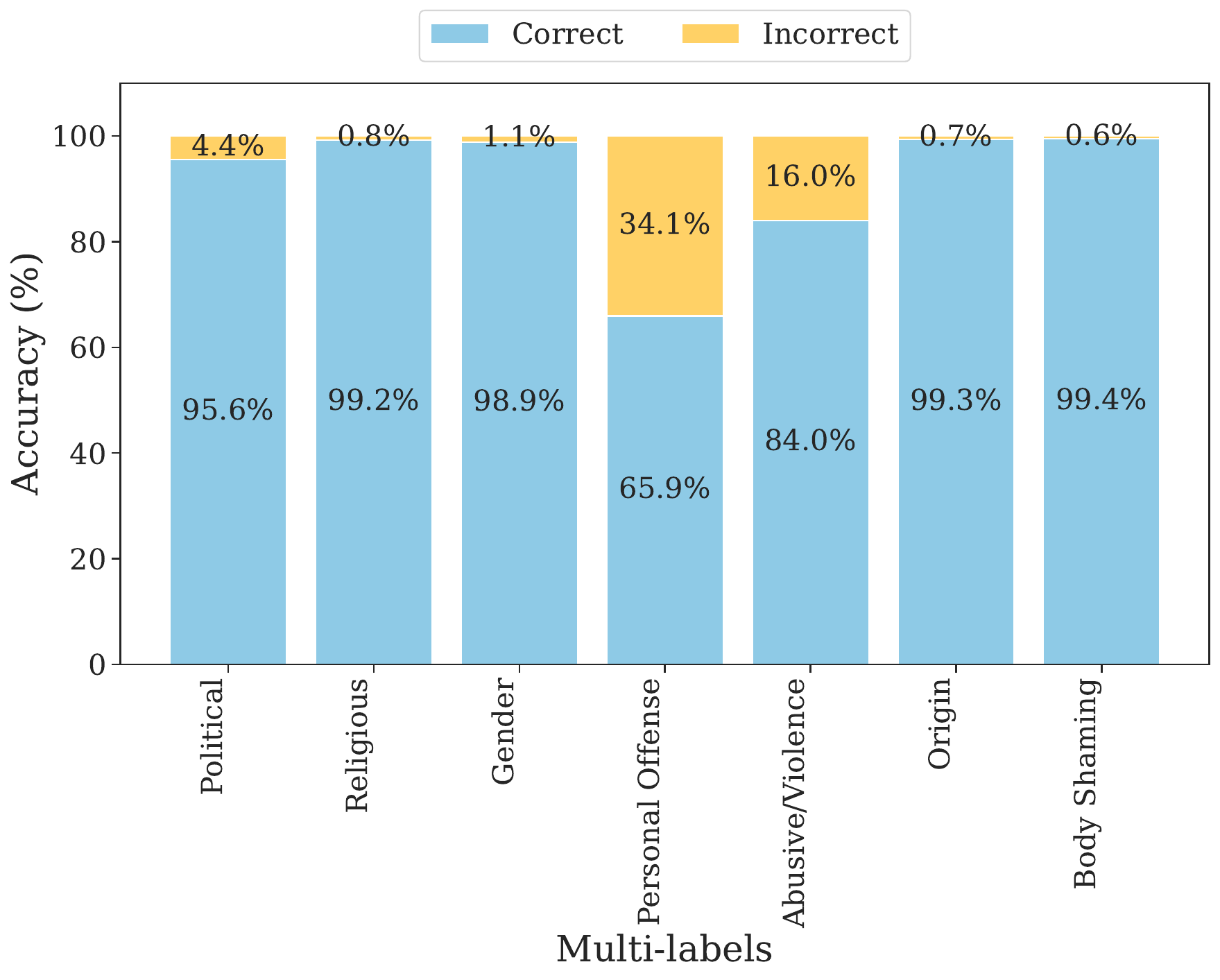}
    \caption{Category-wise macro-F1 score of the best-performing model on \textsc{BanTH}.}
    \label{fig:multi-label-accuracy}
\end{figure}

\section{Error Analysis}

\subsection{High Error in Personal Offense}
 
 Figure \ref{fig:multi-label-accuracy} illustrates per-label accuracy with the best performing model, TB-mBERT. Most hate samples are labeled under \textit{Personal Offense} (Tab. \ref{tab:video-categorywise-multilabel}). Interestingly, \textit{Personal Offense} has the lowest macro-F1 score for mBERT (Fig. \ref{fig:multi-label-accuracy}). This is due to \textit{Personal Offense} exhibiting higher variations of transliterated text (Fig. \ref{fig:bangla_wordcloud},\ref{fig:english_wordcloud}), making it difficult to attribute specific words that signify this category. The higher co-occurrence of this category with other categories might also contribute to the higher error.

\subsection{Representative Words}
Analyzing the word cloud in Fig. \ref{fig:bangla_wordcloud},\ref{fig:english_wordcloud}, we observe that certain words representative of their respective hate categories frequently occurs in the hate categories with higher performance. For instance, \textit{Muslim} frequently appears in "Religious" hate, mohila (woman) and meye (girl) in "Gender" hate, and so on. In contrast, the frequently occurring words for "Personal Offense" appear to be generic terms, e.g. dalal (broker), police, and tui (you). Consequently, it becomes more difficult for language models to classify a word as "Personal Offense" due to the lack of representative words.


\begin{figure}[t]
    \centering
    \includegraphics[width=.45\textwidth]{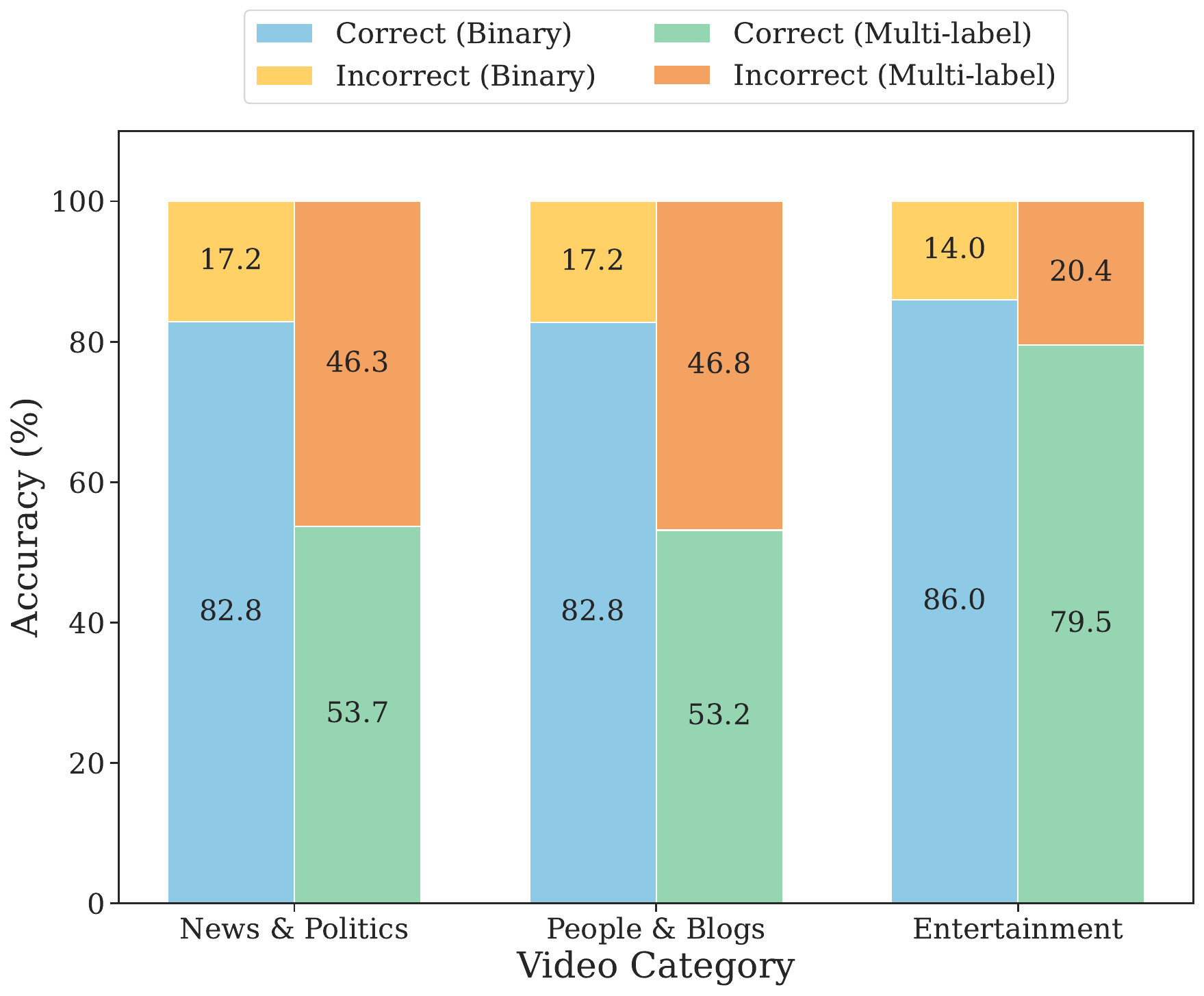}
    \caption{Video category-wise class classification accuracy on the BanTH dataset with the best-performing model. We consider Macro-Accuracy for Binary and Subset-Accuracy for Multi-label Classification.}
    \label{fig:channelwise-accuracy}
\end{figure}

\begin{table*}[h!]
    \centering
    \resizebox{\textwidth}{!}{
    \begin{tabular}{p{4cm}p{4cm}p{0.75cm}p{3cm}p{4cm}p{4cm}}
    \toprule
    
    \textbf{Transliterated Text}  & \textbf{English Translation} & \textbf{Pred.}& \makecell[c]{ \textbf{Reason by}\\ \textbf{GPT-4o+E}}& \makecell[c]{ \textbf{Reason by}\\ \textbf{GPT-4o+E+T}}  & \makecell[c]{ \textbf{Translation by}\\ \textbf{GPT-4o+E+T}}\\
    \midrule
    world er shob deshei torun shomaj montri mp hoy khali Bangladesh chara eitar karon holo onar moto kichu political peraon jara sharajibon nijei kheye jete chay manusher jonno korte chay na&	In all the countries of the world, youths become social ministers and MPs, but except for Bangladesh. The reason behind this is a kind of political anarchy those who live their entire lives for themselves not for the welfare of people.	&\textcolor{darkgreen}{NH}&	The text criticizes the entire political generation with derogatory implications. & The text criticizes political leaders in Bangladesh for not focusing on youth leadership, but it does not incite hatred or use dehumanizing language.&In all countries of the world, young people are made ministers or MPs, except in Bangladesh. The reason for this is that some political persons want to selfishly continue benefiting themselves instead of working for the people.\\
    \midrule
    Tomrao to bikri hoe geco& You have been sold too	&\textcolor{darkgreen}{NH}&	The text implies corruption and selling out, suggesting negative stereotyping.&	The phrase does not express hatred or discrimination towards any specific group or individual.&	They have also been sold out.\\
    \midrule
    Chup catukar khuni&	Shut up killer sycophant &\textcolor{darkgreen}{H}&	The text seems to express frustration but lacks direct hate indicators.&	The phrase 'chup catukar khuni' implies killing, which incites violence and hate.&	Shut up, murderous sycophant.\\
    \midrule
    Ore dhora hok ainer awtai ana hok&	Let him be caught and brought under justice	&\textcolor{darkgreen}{NH}&	The text calls for arrest and punishment. & The text talks about bringing someone under the law without expressing or inciting hate or violence.	& Let him be caught and brought under the law.\\
    \midrule
    Agey apnader mordhe indian  gulare khuje ber koren& First, find the Indians among yourselves	&\textcolor{darkgreen}{H}	&	The text calls for identifying certain individuals without expressing hate.&	The text stereotypes and discriminates based on origin, which is a hate indicator.	& First, find the Indian infiltrators among yourselves.\\
    

    
    \midrule
    
    Myb eta India cal parar meye & Maybe she is a girl from some slum in India. & \textcolor{red}{H} & The text derogatorily implies an Indian woman based on location, with a stereotype & There is no evident expression of hate speech in identifying someone by a phrase or as a resident of a place.& Maybe she is a girl from India Calcutta.\\
    \midrule
    eto jhamelar ki orkar era beche thakle sob kiso vole abaro chori bodmashi korbe & What’s the use of so much trouble? If they stay alive, they will forget everything and steal and do mischief again. & \textcolor{red}{NH} & The text discusses obstacles without expressing hate. & The text suggests that a group of people, if they remain alive, will engage in theft and misconduct again. It stereotypes and implies inherent criminal behavior. & What is the point of all this trouble, if they stay alive, they will forget everything and steal and commit mischief again.\\
    \midrule
    Tora nijerai to political party er pasa chatis. & You yourselves are the sycophants of political parties. &\textcolor{red}{H} & The text accuses others of opportunistic political affiliations.& The text is a general statement about political affiliation without direct hate speech indicators. It appears accusatory but not incite violence or hatred.& You people yourselves are involved with the political party.\\
    \midrule
    Bhai ja korso tai paiso..tmra joto marso oitar hisab k dibe ?? & Brother, you got what you deserved... will you give an account for how much you have killed? & \textcolor{red}{NH} & The text discusses the consequences of actions without hate indicators. & This text contains implicit threat and justification for physical violence, possibly in a retaliatory context. & Brother, you got what you did.. Who's going to account for how much you hit?\\
    \midrule
    Na bnp na awami league Islami Andolon Bangladesh best.& Neither BNP nor Awami League, Islami Andolon Bangladesh is the best. & \textcolor{red}{NH} & The text expresses political preference without any hate indicators. &This text promotes political supremacy, suggesting one political movement is superior to others.& No BNP, no Awami League, Islami Andolon Bangladesh is the best.\\
    \bottomrule
    \end{tabular}}
    \caption{Comparison of predictions and reasoning generated by GPT-4o using Explanation (E) and Translation (T) prompting. \textcolor{darkgreen}{Green} indicates that the prediction (H:Hate and NH:Non-Hate) was correctly labeled by GPT-4o+E+T but incorrectly labeled by GPT-4o+E, and \textcolor{red}{red} indicates incorrect prediction by both the models.}
    \label{tab:no-match-comparison}
\end{table*}

\subsection{Translation by LLMs}
As observed in Tab. \ref{tab:no-match-comparison}, LLMs can naturally translate Bangla transliterated text into English, even without explicit instructions. However, when explicitly instructed to translate, the LLM tends to give more literal translations. We also observe that erroneous translations often lead to incorrect classifications, especially with regional dialects or terms lacking direct English equivalents. English translations are generally observed to outperform Bangla ones. When not prompted to translate, the LLM focuses on identifying hate-related terms, which works for explicit hate speech but struggles with subtler forms of offensive content, leading to misclassification.

\section{Related Work} \label{sec:rw}

\noindent\textbf{Hate Speech.}~~Natural Language Processing (NLP) tasks on hate speech primarily involve hate speech detection \cite{schmidt-wiegand-2017-survey} which can be binary \cite{mutanga2020hate,bose-su-2022-deep}, multi-class \cite{yigezu2023transformer,hashmi2024multi}, or multi-label classification \cite{mollas2022ethos}. Multi-class and multi-label tasks generally categorize hate based on race, religion, gender, origin, and disability. Hate speech detection has also been explored in multi-modal \cite{gomez2020exploring,kiela2020hateful} and multi-lingual \cite{aluru2020deep,ousidhoum-etal-2019-multilingual} settings.
\\
\\
\noindent\textbf{Bangla Hate Speech.}~~As a low-resource language, Bangla has limited work on hate speech detection \cite{romim-etal-2022-bd,das-etal-2022-hate-speech} but has made significant progress on other hate-related tasks involving abusive content \cite{hussain2018approach,jahan2022banglahatebert}, cyberbullying \cite{emon2022detection}, sexism \cite{jahan2023deep}, and violence \cite{fahim-2023-aambela}. The multi-label variants are generally categorized based on religion, politics, and gender \cite{sharif2022m}. Key areas of multi-labeling tasks include aggression \cite{hossain2023multilabel}, cyberbullying \cite{saifuddin2023enhancing}, hate speech \cite{shakil2022hybrid, math12132123}, and toxic content \cite{belal2023interpretable}.
\\
\\
\noindent\textbf{Transliterated Bangla.}~~~Research on NLP tasks using transliterated or romanized Bangla is notably scarce, with limited works on back-transliteration \cite{shibli2023automatic}, abusive content detection \cite{jahan2019abusive,sazzed2021abusive}, hate speech detection \cite{das-etal-2022-hate-speech}, offensive language identification \cite{raihan2023offensive}, sentiment analysis \cite{hassan2016sentiment}, and cyberbullying detection \cite{ahmed2021deployment}. While multi-label classification tasks on transliterated Bangla have been explored as sentiment and emotion analysis tasks \cite{tripto2018detecting}, there remains a significant research gap on multi-label hate speech classification for transliterated Bangla.

\section{Conclusion}
Our novel multi-label hate speech dataset, \textsc{BanTH}, addresses a crucial research gap in the domain of transliterated Bangla. Extensive experimentations on both traditional language models and LLMs highlight their usefulness in their unique settings. We envision our novel prompting strategies to be generalized to the processing of transliterated text in other languages. We believe our dataset can serve as a valuable resource in the creation of safer digital platforms for the future.



\section*{Limitations}
\label{sec:limitations}

The \textsc{BanTH} dataset contains 27167 non-hate samples out of 37350, i.e. a dummy classifier predicting all samples as non-hate achieves an accuracy of 72.73\%. This undermines the performance of a few methods, notably, GPT-3.5 performs worse than this baseline. 

Our work does not cover all possible hate classes and may not adequately represent other regionally relevant categories, e.g. hate based on socio-economic conditions. As the dataset has been sourced from YouTube comments, there are insufficient instances of such categories. Adding a low proportion of these samples will create more imbalance in multi-label classification. 

We emphasize that the inter-annotator score is comparatively lower than in other classification tasks, primarily due to the inherently subjective nature of hate speech and the variations in transliteration. The expert agreements have also been found lower than the normal annotators, which can be attributed to experts having higher levels of disagreement in defining hate categories due to their deeper understanding of the gray areas in interpretation. In contrast, the simpler and straightforward thinking can be attributed to the higher agreement scores between general annotators.



\section*{Ethical Considerations}
The use of comments scraped from the YouTube videos complies with the YouTube API's terms of services\footnote{\url{https://developers.google.com/youtube/terms/api-services-terms-of-service}}. All forms of Personal Identification Information (PII) have been removed from the dataset to prevent privacy violations. We ensured fair distribution of annotation workload. The hired annotators and domain experts were compensated on an hourly basis at a rate above the industry standard.

\section*{Acknowledgments} We sincerely appreciate the generous support of our project sponsor, Penta Global Limited, Bangladesh.

\bibliography{main}

\appendix
\section{Additional Related Work} \label{sec:arw}

\noindent\textbf{Hate-like Tasks.}~~Hate speech lacks a consistent definition across the literature and has significant overlap with similar detection tasks involving offensive language \cite{pradhan2020review,ranasinghe-zampieri-2020-multilingual}, sexism \cite{chiril-etal-2021-nice-wife,shifat2024penta}, abusive content \cite{nobata2016abusive}, slang \cite{pei-etal-2019-slang}, and cyberbullying \cite{verma-etal-2022-benchmarking,ahmed2022performance}. The studies on hate speech also focused on areas, such as explainability \cite{mathew2021hatexplain,fahimhatexplain},  fairness \cite{mostafazadeh-davani-etal-2021-improving}, and bias mitigation \cite{xia-etal-2020-demoting}. 
\\
\\
\noindent\textbf{Non-English Hate Speech.}~~While the majority of the hate speech datasets are in US English \cite{tonneau-etal-2024-languages}, the task has been explored in several non-English languages including Arabic \cite{anezi2022arabic}, Hindi \cite{das-etal-2022-hatecheckhin}, Indonesian \cite{ibrohim-budi-2019-multi}, Korean \cite{kang2022korean,lee-etal-2022-k}, Marathi \cite{patil-etal-2022-l3cube}, and Spanish \cite{plaza2021comparing}. Due to the prevalence of English alphabets in digital spaces, some of these works involve transliterated \cite{hettiarachchi2020detecting} and code-mixed texts \cite{bohra-etal-2018-dataset}.

\section{BanTH Dataset Annotation Guidelines}
\label{sec:annotation-guidelines}
This document outlines the procedures required for accurate annotation of texts within the BanTH dataset. Annotators are expected to follow each step meticulously to ensure consistency and precision. Your role as an annotator is critical to maintaining the integrity and quality of the dataset.
\paragraph{Binary Classification - Hate/Non-Hate: }Texts containing offensive, abusive, or harmful language targeting individuals or groups based on their identity, characteristics, or affiliations are to be labeled hate. If the text does not include offensive or harmful content it is to be labeled non-hate.
    \paragraph{Steps:}
    \begin{enumerate}
        \item Review the text in transliterated Bangla.
        \item Classify it as either Hate Speech or Non-Hate Speech.
        \item If the comment is Non-Hate, stop here. No further labeling is necessary.
    \end{enumerate}
\paragraph{Multi-Label Hate Speech Classification:}
For texts classified as Hate Speech, assign one or more relevant labels from the following categories. Each text may fit multiple categories, so assign all that apply.
    \begin{itemize}
    \item {\textbf{Political:} Texts that attack, criticize, or promote hate based on political affiliations, views, or ideologies. This includes language dehumanizing political opponents or encouraging harm.\\
    \textbf{Example:}\\
    \textbf{[BN]} "Oder rajniti desh ta noshto korche."\\
    \textbf{[EN]} "Their politics are destroying the country."}
    \item {\textbf{Religious:} Texts targeting someone’s religious beliefs or lack thereof, or that promote violence or discrimination based on religion.\\
    \textbf{Example:}\\
    \textbf{[BN]} "Ei dhormo manushkei bhag kore."\\
    \textbf{[EN]} "This religion only divides people."}
    \item {\textbf{Gender:} Texts that reinforce harmful gender stereotypes or discriminate based on gender identity or expression. Includes sexist or transphobic remarks.\\
    \textbf{Example:}\\
    \textbf{[BN]} "Meyera kaj korte pare na, barir kaji kore."\\
    \textbf{[EN]} "Women can’t work, they only do household chores."}
    \item {\textbf{Personal Offense:} Deeply insulting or offensive remarks directed at an individual, which do not target a specific group but are highly personal.\\
    \textbf{Example:}\\
    \textbf{[BN]} "Tui ekta boka chele."\\
    \textbf{[EN]} "You’re a stupid boy."}
    \item {\textbf{Abusive/Violence:} Texts that contain explicit threats of violence or encourage violent acts against individuals or groups.\\
    \textbf{Example:}\\
    \textbf{[BN]} "Tor matha guriye debo."\\
    \textbf{[EN]} "I’ll smash your head."}
    \item {\textbf{Origin:} Texts targeting someone based on their nationality, ethnicity, or race. This includes racial slurs or calls for exclusion based on origin.\\
    \textbf{Example:}\\
    \textbf{[BN]} "Ei desher manush er kono kotha nai."\\
    \textbf{[EN]} "People from this country don’t matter."}
    \item {\textbf{Body Shaming:} Texts that criticize or mock someone’s physical appearance, such as body size or traits, including physical disabilities or conditions.\\
    \textbf{Example:}\\
    \textbf{[BN]} "Tor pet eto boro kano?"\\
    \textbf{[EN]} "Why is your stomach so big?"}
    \end{itemize}
Each sample will be annotated by three different annotators. After all of this, domain experts will verify the labels to ensure quality.\\

\noindent\textbf{Notes:} Firstly, carefully read and understand each text before assigning labels. Note that, a sample can receive multiple labels if it fits into more than one category. Please ensure cross-verification by collaborating with fellow annotators. If the problem is not solved even after collaboration then contact domain experts because accuracy is essential for a high-quality dataset.

\section{Additional Details of \textsc{BanTH}}

\subsection{Formal Description of Target Groups} \label{app:target_descriptopn}
The description of the target group with suitable examples is given in the following:\\

   \noindent \textbf{Political:}  Statements that marginalize, threaten, or incite violence against individuals or groups based on their political affiliations or ideologies. Includes calls for harming political opponents, dehumanizing metaphors applied to political groups, or the deliberate spread of disinformation intended to foment hatred.\\ 
   
    \noindent \textbf{Religious:} Expressions targeting individuals or groups due to their religious beliefs, practices, or lack thereof. Includes language that demonizes religious communities, advocates for discrimination against religious groups, or incites violence toward religious institutions or adherents.\\

    
    \noindent \textbf{Gender:} Extends beyond sexist remarks and includes language promoting gender-based violence, discrimination based on gender identity, misogynistic or misandrist content, transphobic speech, and language reinforcing harmful gender stereotypes. \\

    
    \noindent \textbf{Personal Offense:} Deeply insulting or offensive language directed at an individual. Includes mocking tragedies, using insulting names, or making derogatory comments.\\

    
    \noindent \textbf{Abusive/Violence:} Explicit threats of violence, detailed descriptions of harm one wishes to inflict on others, or speech that glorifies or encourages violent acts against individuals or groups.\\ 

    
    \noindent \textbf{Origin:} Expressions that target individuals or groups based on their national, ethnic, or racial background. Includes racial slurs, promotion of racist/nationalist ideologies, calls for racial segregation or deportation, or speech that attributes negative characteristics to groups.\\ 

    
    \noindent \textbf{Body Shaming:} Derogatory comments on a person's physical appearance, often related to weight, shape, or size. Can extend to mocking people with physical disabilities, skin conditions, or other visible physical traits.\\  





\subsection{Wordcloud}
\label{sec:wordcloud}

\begin{figure*}[h!]
    \centering
    \subfigure[Political]{
        \includegraphics[width=0.48\textwidth]{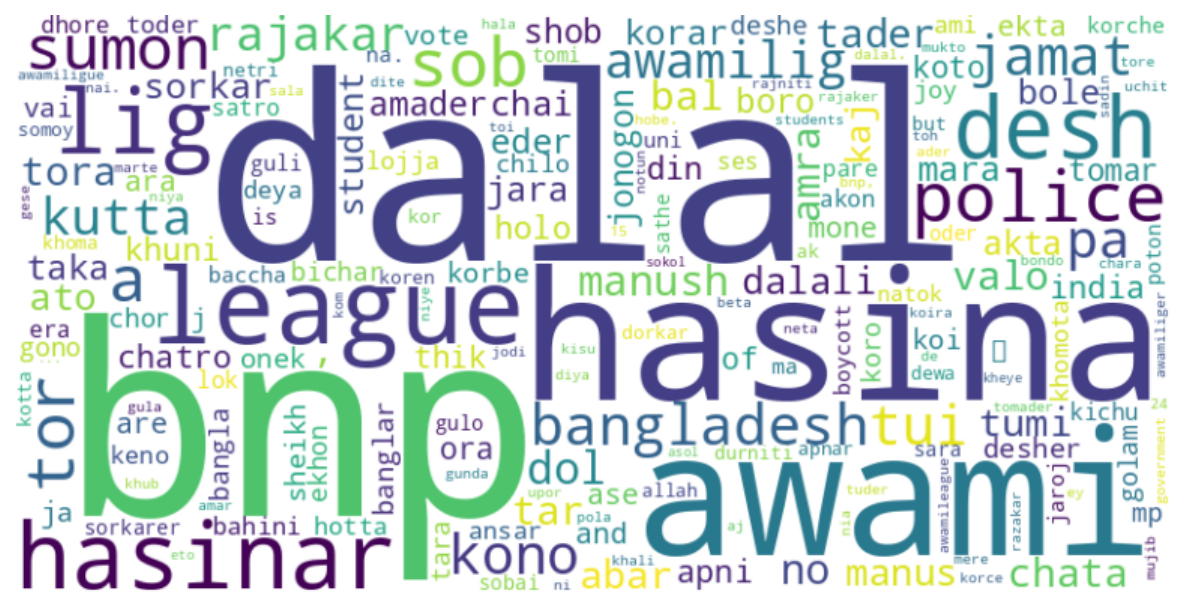}
    }
    \subfigure[Religious]{
        \includegraphics[width=0.48\textwidth]{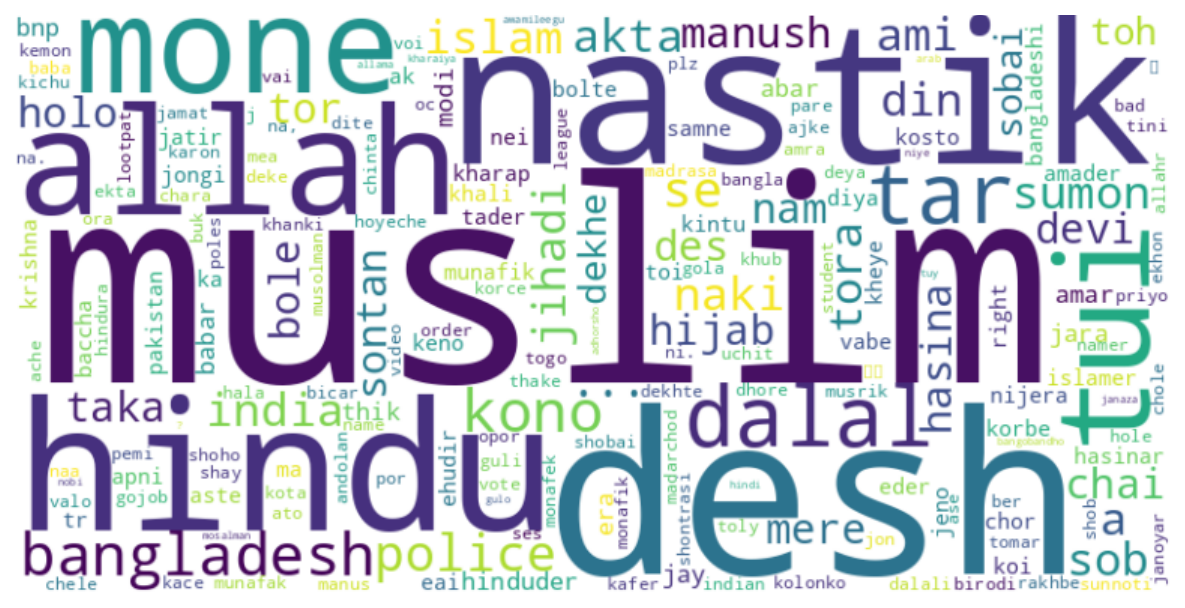}
    }
    \subfigure[Gender]{
        \includegraphics[width=0.48\textwidth]{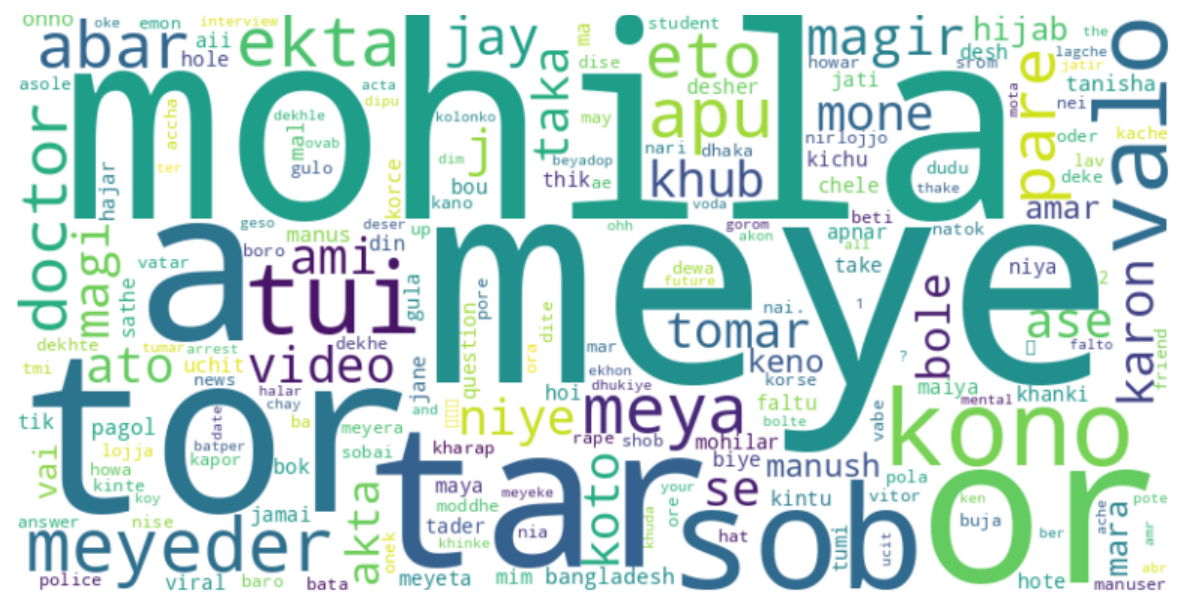}
    }
    \subfigure[Personal Offense]{
        \includegraphics[width=0.48\textwidth]{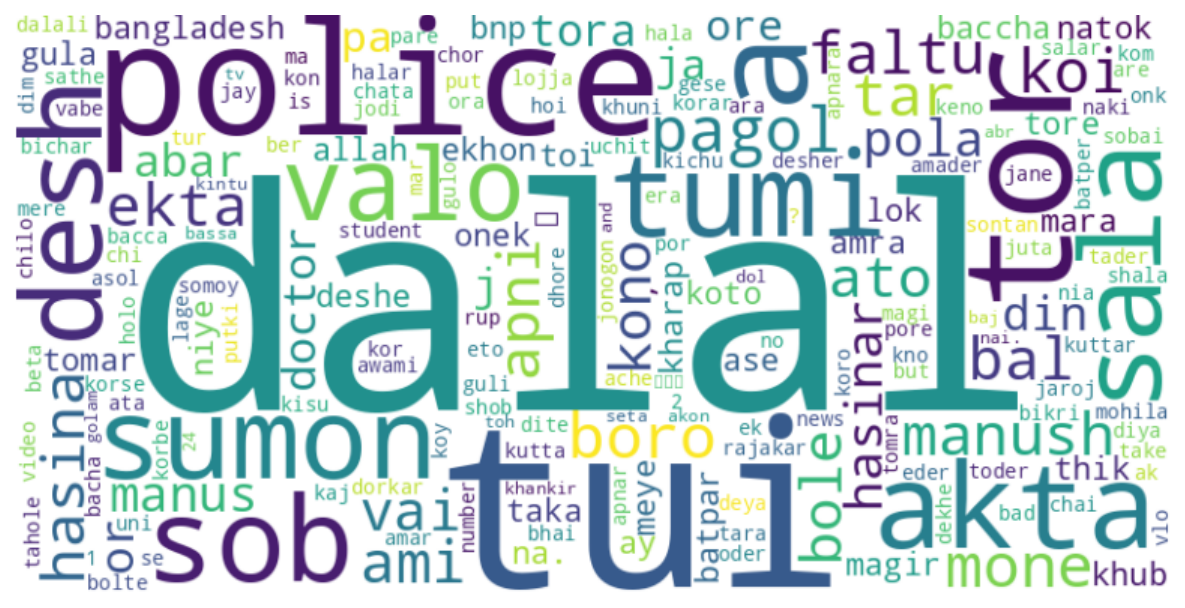}
    }
    \subfigure[Abusive/Violence]{
        \includegraphics[width=0.48\textwidth]{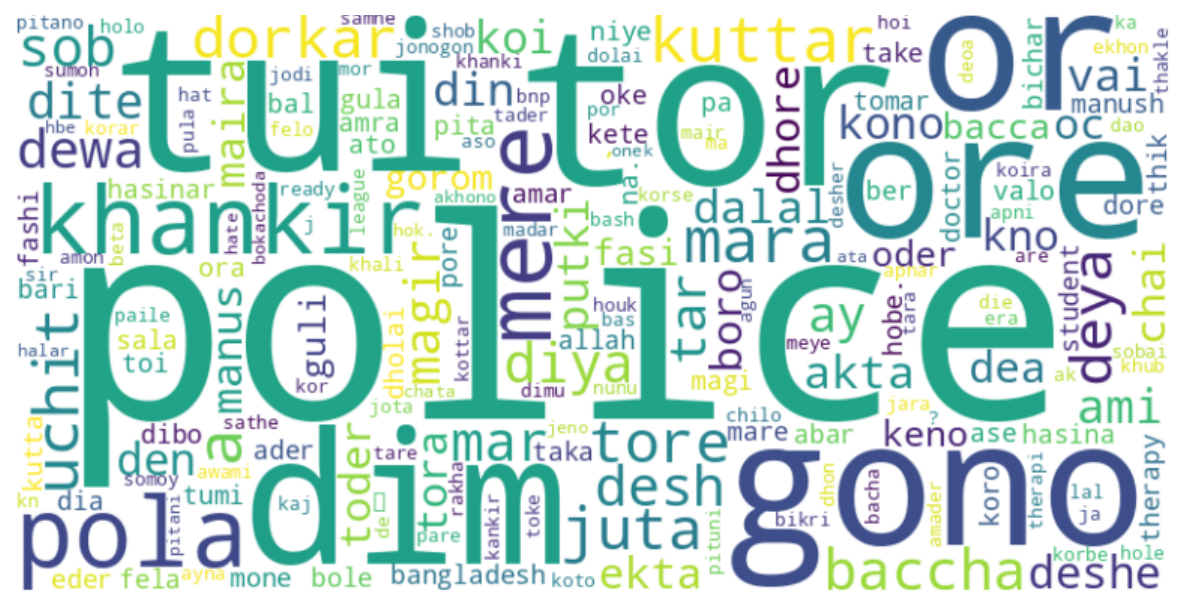}
    }
    \subfigure[Origin]{
        \includegraphics[width=0.48\textwidth]{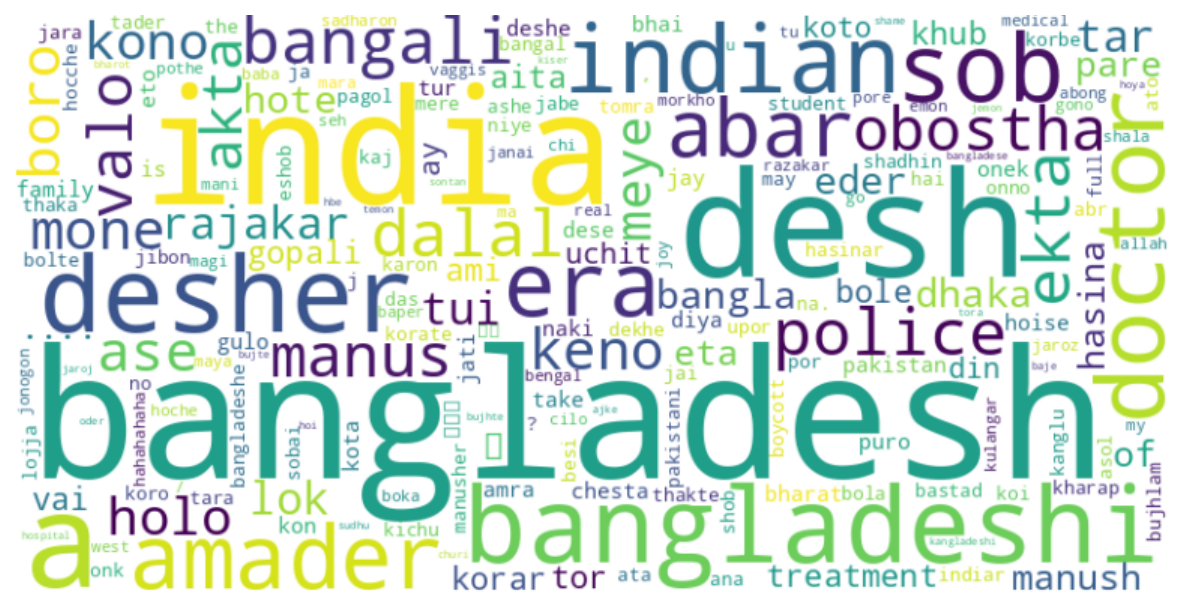}
    }
      \subfigure[Body Shaming]{
        \includegraphics[width=0.48\textwidth]{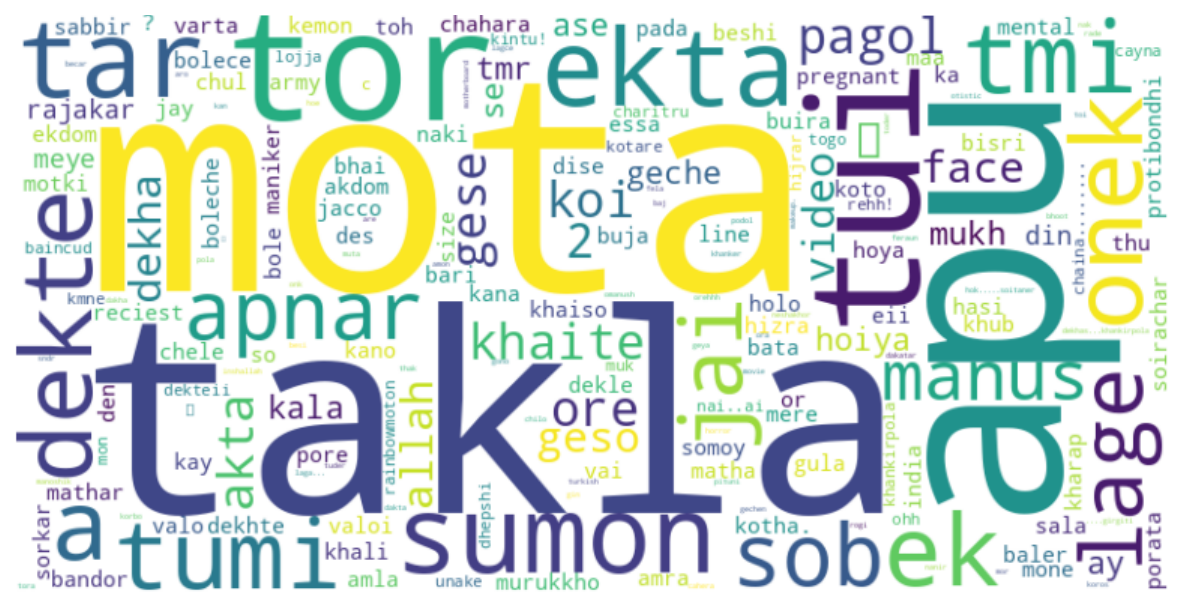}
    }
    \caption{Word clouds constructed from transliterated Bangla texts in the \textsc{BanTH} dataset for each label.}
    \label{fig:bangla_wordcloud}
\end{figure*}

\begin{figure*}[h!]
    \centering
    \subfigure[English Political]{
        \includegraphics[width=0.48\textwidth]{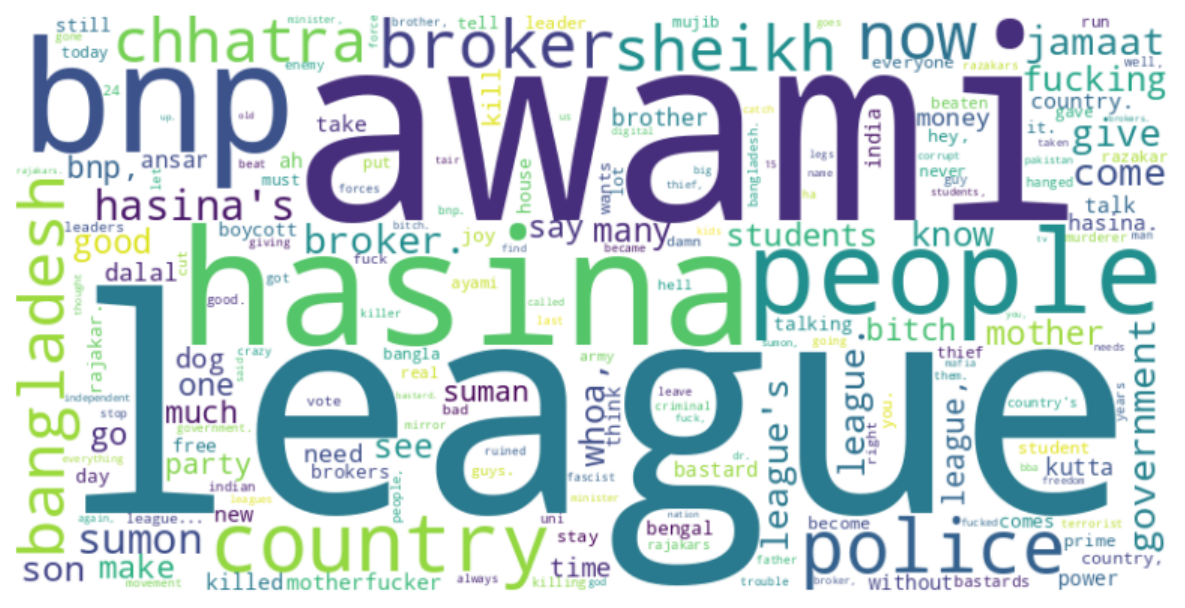}
    }
    \subfigure[Religious(English Translated)]{
        \includegraphics[width=0.48\textwidth]{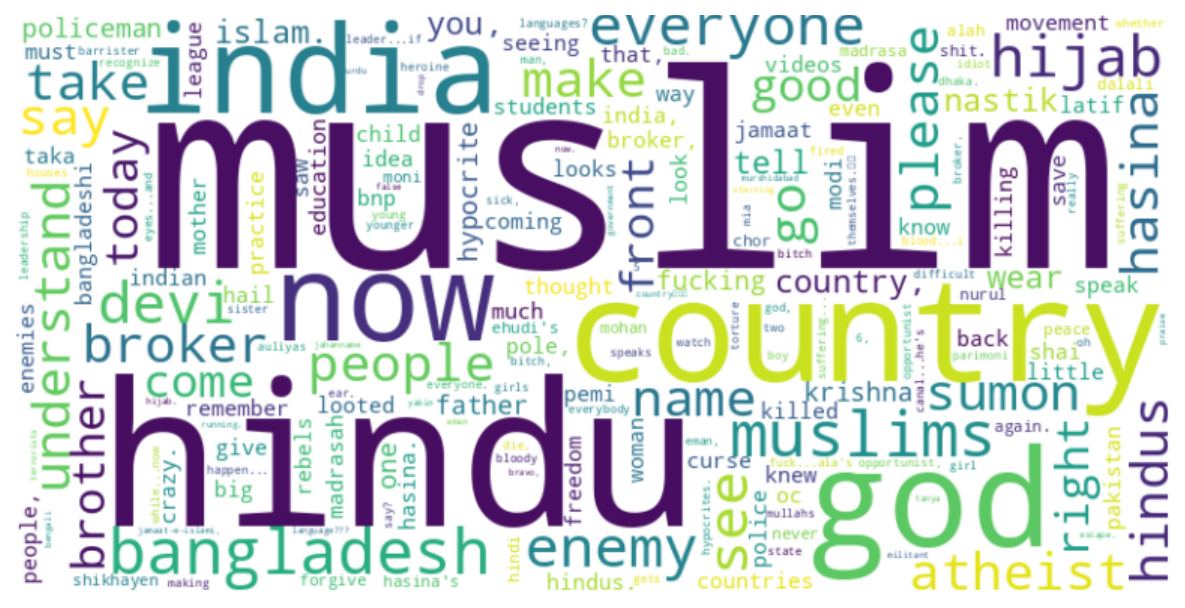}
    }
    \subfigure[Gender (English Translated)]{
        \includegraphics[width=0.48\textwidth]{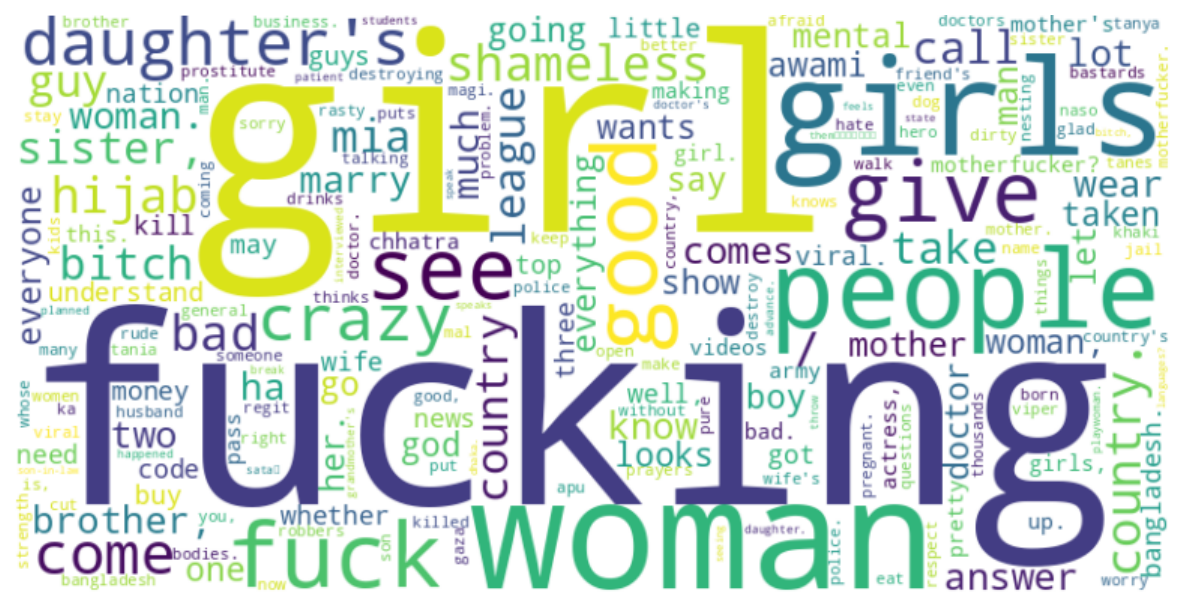}
    }
    \subfigure[Personal Offense (English Translated)]{
        \includegraphics[width=0.48\textwidth]{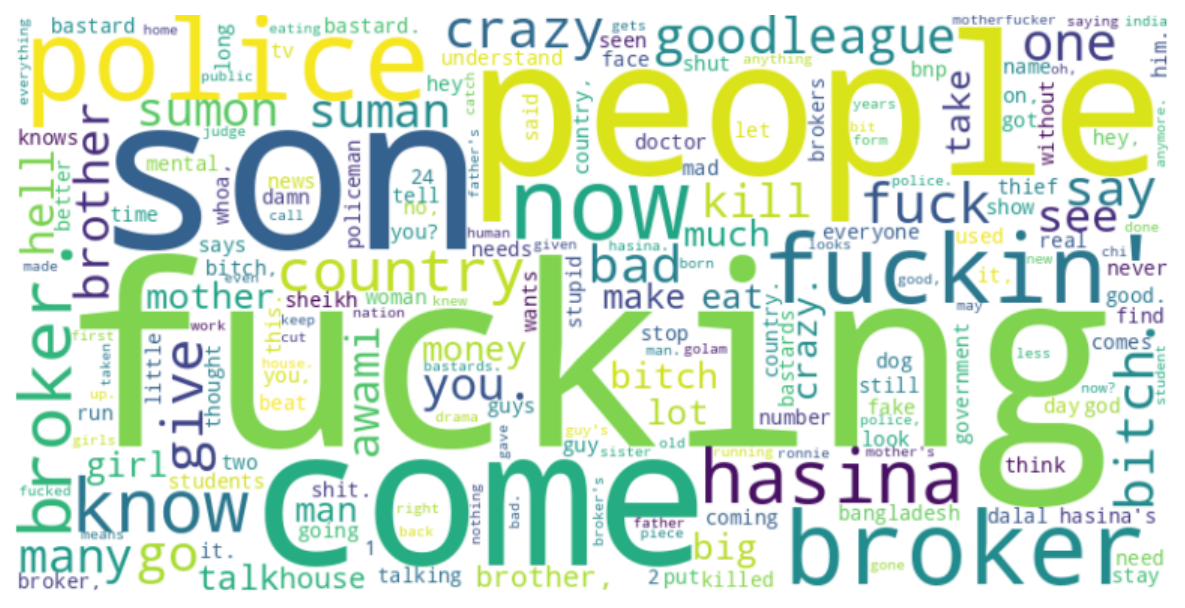}
    }
    \subfigure[Abusive/Violence (English Translated)]{
        \includegraphics[width=0.48\textwidth]{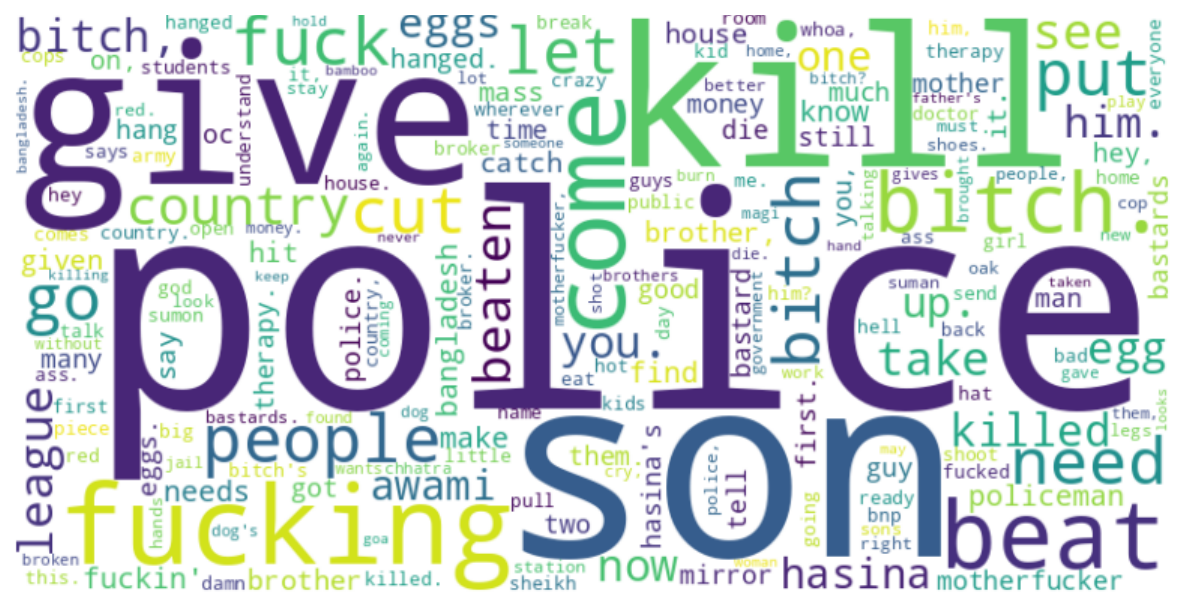}
    }
    \subfigure[Origin (English Translated)]{
        \includegraphics[width=0.48\textwidth]{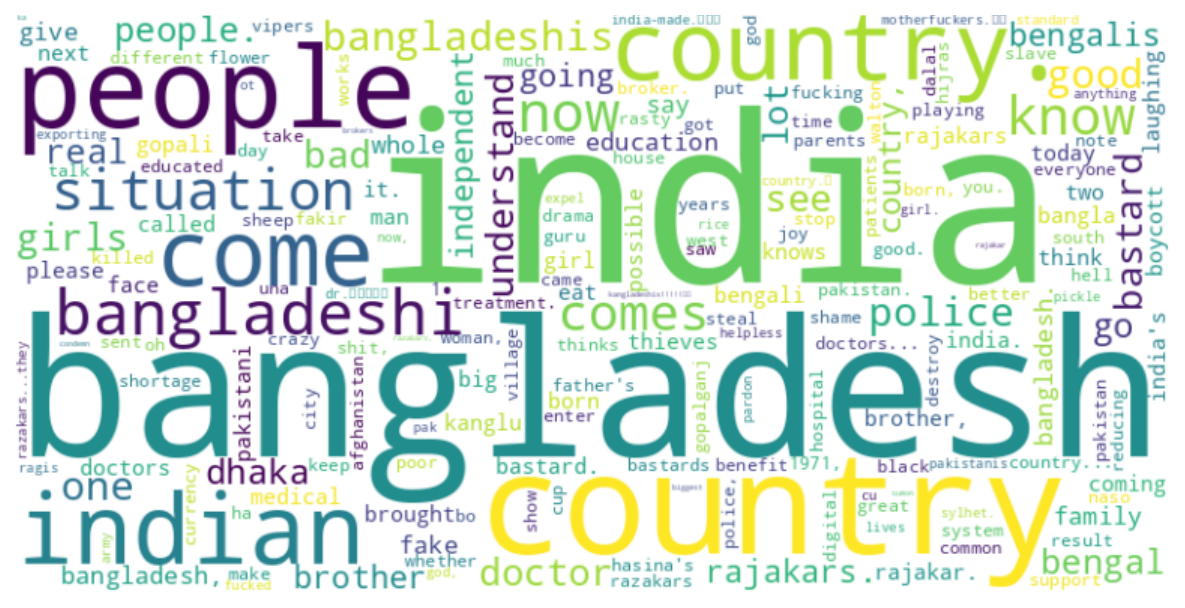}
    }
    \subfigure[Body Shaming (English Translated)]{
        \includegraphics[width=0.48\textwidth]{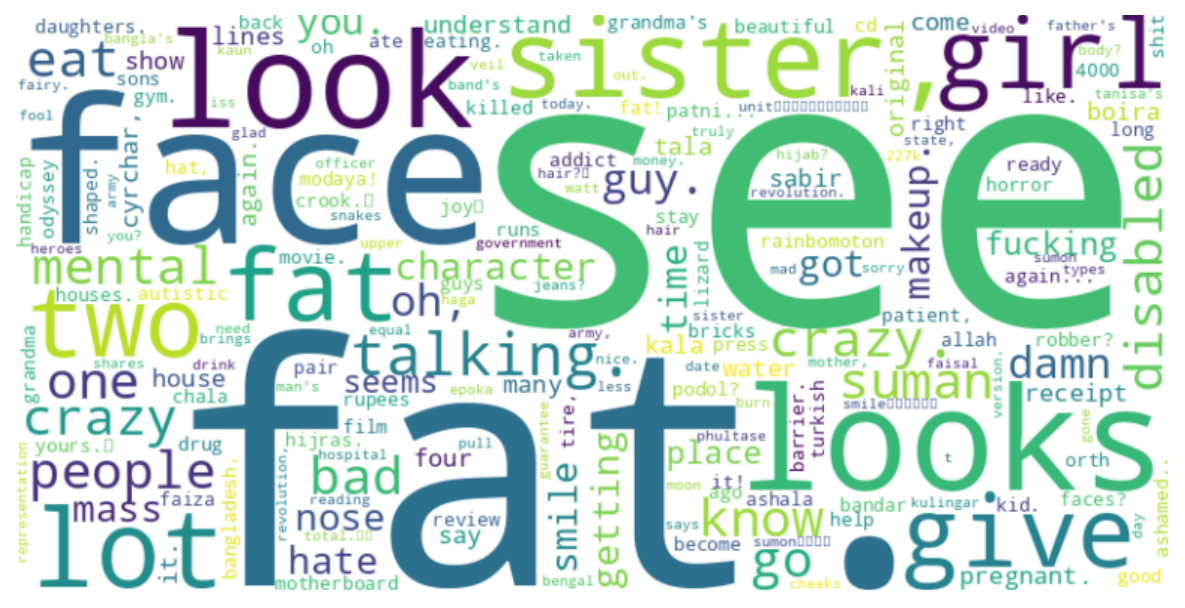}
    }
    \caption{Word clouds constructed from English translations of the Bangla texts in the \textsc{BanTH} dataset for each label.}
    \label{fig:english_wordcloud}
\end{figure*}

From Figure \ref{fig:bangla_wordcloud} and the corresponding English translations in Figure \ref{fig:english_wordcloud}, we can visualize the mostly mentioned words for each label in the BanTH dataset. For \textit{Political} label, people often referred to Hasina (ex-prime minister of Bangladesh), bnp (a major political party in Bangladesh), league/lig (wing of a political party in Bangladesh), dalal (broker), police (a law enforcement force), etc. As we have covered transliterated Bangla YouTube comments of July 2024, the BanTH dataset includes the public responses related to the political unrest of Bangladesh during that period. 

In \textit{Religious} labeled sample texts, people often mention Muslim, Hindu nastik (atheist), Allah (the one and only God in Islam), etc. \textit{Gender} related hate comments are mostly related to females. In \textit{Personal Offense} case, commenters address dalal (broker) and police (a law enforcement force) contemptuously using tui/tor. The same case is evident in \textit{Abusive/Violence} labeled samples because the police showed violence during the student protest of July 2024. 

As transliterated Bangla is used by mainly Indian and Bangladeshi people, \textit{Origin} related hate is often directed towards Bangladeshi and Indian people. For \textit{Body Shaming} classified samples, we see people refer to others as mota (fat) and takla (bald) to body shame and it is directed mostly towards females as apu (sister) is one of the most used words. Although the BanTH dataset is concentrated on these types of hateful words, it may not be the case when the source is different.

\section{Experimental Setup} \label{app:exp_setup}

For Language Models (LMs) baseline experiments, we fine-tuned a range of pre-trained LMs to establish baselines for binary and multi-label classification on the BanTH dataset. The models include multilingual architectures like XLM-RoBERTa\cite{xlm-roberta} and mBERT\cite{mbert}, as well as language-specific models such as BanglaBERT\cite{banglabert} and BanglishBERT\cite{banglabert}, which are designed for Bangla and code-mixed transliterated Bangla text, respectively. We also employed BanglaHateBERT\cite{banglahatebert}, specifically optimized for detecting hate speech in Bangla, along with IndicBERT\cite{indicbert} and MuRIL\cite{muril}, tailored for Indian languages. Additionally, we included models like CharBERT\cite{charbert}, and VAC-BERT\cite{vacbert} to capture diverse model architectures, including lightweight models and character-level representations. 

The models were fine-tuned with a learning rate of 2e-5, using the Adam optimizer. The maximum sequence length was set to 512 tokens, and training was performed for 5 epochs with a batch size of 32. Early stopping was applied based on the validation loss to prevent overfitting. For prompt-based experiments, we used OpenAI API to get LLM responses for GPT-3.5 and GPT-4o models. For measuring the performance, we consider \texttt{Macro-F1} \& \texttt{Accuracy} metrics for binary classification and \texttt{Macro-F1}, \texttt{Subset-Accuracy} \& \texttt{Hamming Loss} for multilabel classification. In our further pretraining experiments, we randomly masked 15\% of the tokens and used Masked Language Model loss (MLM loss) as the pretraining objectives. We used a learning rate of 1e-5, with batch\_size = 32 and we trained the the TB-Encoder models for 5 epochs on the whole training dataset.

\section{Performance Metrics} \label{app:performance_metrics}


\textbf{F1-Score.}~~~Harmonic mean of precision and recall, providing a balance between the two. It is particularly useful in cases of imbalanced datasets, as it accounts for both false positives and false negatives. For multi-label classification, the F1-Score is computed per label, and the final score is obtained by averaging across all labels.\\

\noindent\textbf{Macro-Accuracy.}~~~For binary classification, we report macro-accuracy, which averages the accuracy across both classes (positive and negative), providing a balanced view of the model's performance across the two categories. This is especially relevant when dealing with class imbalance, as it ensures that both classes are equally represented in the final accuracy measure.\\

\noindent\textbf{Subset-Accuracy.}~~~Requires an exact match between the predicted and true label sets for each instance, meaning that all labels must be predicted correctly for a prediction to be considered accurate \cite{multilabel}.\\

\noindent\textbf{Hamming Loss} Measures the fraction of labels that are incorrectly predicted, either by being wrongly assigned or missed altogether. This metric is useful for multi-label classification as it accounts for both false positives and negatives across all labels. A lower Hamming Loss value indicates better performance as it signifies fewer labeling mistakes. \\

\definecolor{darkgreen}{RGB}{0,100,0}
\clearpage

\section{Prompts for Classification on BanTH dataset}
\label{sec: prompts}
\begin{figure}[h!]
    \centering
\begin{tcolorbox}[colback=gray!5!white, colframe=black, title= Base Prompt for Binary Classification (0 shot), width=0.47\textwidth]
\small
You are an expert at detecting hate speech in Bangla text (written in Latin script/English letters). Your task is to classify texts as either hate speech (true) or non-hate speech (false).\\

HATE SPEECH DEFINITION:\\
Text that expresses or incites hatred, discrimination, or violence against individuals or groups based on:\\
- Political affiliation \\ 
- Religion (e.g., Hindu, Muslim, Buddhist, Christian)\\
- Gender\\
- Personal offense\\
- Abusive or violence\\
- Origin\\
- Body Shaming\\

KEY INDICATORS OF HATE SPEECH:\\
1. Dehumanizing language or comparisons\\
2. Calls for violence or harm\\
3. Discriminatory slurs or epithets\\
4. Stereotyping entire groups negatively\\
5. Promoting supremacy of one group over others\\

LABELING INSTRUCTIONS:\\
1. For each text, determine if it contains hate speech indicators\\
2. If any hate indicators are present, classify as hate speech (true)\\
3. If no hate indicators are present, classify as non-hate speech (false)\\

SPECIAL CONSIDERATIONS:\\
- Consider local context and cultural references\\
- Be mindful of dialectal variations in spelling\\
- Pay attention to code-mixing (Bangla-English hybrid phrases)\\
- For ambiguous cases, focus on the presence of hate indicators\\

You will receive an array of objects, each containing an 'id' and 'text'. Analyze each text and provide the required classification.
\label{prompt:binary}
\end{tcolorbox}
\end{figure}
\begin{figure}[h!]
    \centering
\begin{tcolorbox}[colback=gray!5!white, colframe=black, title= Base Prompt for Multilabel Classification (0 shot), width=0.47\textwidth]
\small
You are an expert at detecting and categorizing hate speech in Bangla text (written in Latin script/English letters). Your task is to classify texts into multiple categories of hate speech.\\

HATE SPEECH CATEGORIES:\\
1. POLITICAL: Targeting individuals or groups based on political affiliation or beliefs\\
2. RELIGIOUS: Targeting individuals or groups based on religion (e.g., Hindu, Muslim, Buddhist, Christian)\\
3. GENDER: Targeting individuals or groups based on gender identity or expression\\
4. PERSONAL OFFENSE: Insults or attacks directed at specific individuals\\
5. ABUSIVE/VIOLENCE: Language that is abusive or incites violence\\
6. BODY SHAMING: Targeting individuals based on physical appearance or body type\\
7. ORIGIN: Targeting individuals or groups based on their place of origin, ethnicity, or nationality\\
8. MISC (Miscellaneous): Other forms of hate speech not covered by the above categories\\

KEY INDICATORS OF HATE SPEECH:\\
1. Dehumanizing language or comparisons\\
2. Calls for violence or harm\\
3. Discriminatory slurs or epithets\\
4. Stereotyping entire groups negatively\\
5. Promoting supremacy of one group over others\\

LABELING INSTRUCTIONS:\\
1. For each text, determine if it contains hate speech indicators for any of the defined categories\\
2. Assign (True) all relevant hate speech categories to the text\\
3. If no hate indicators are present, classify as non-hate speech (empty array for categories)\\

SPECIAL CONSIDERATIONS:\\
- Consider local context and cultural references\\
- Be mindful of dialectal variations in spelling\\
- Pay attention to code-mixing (Bangla-English hybrid phrases)\\
- A single text may belong to multiple hate speech categories\\
- Use the MISC category for hate speech that doesn't fit into other categories\\

You will receive an array of objects, each containing an 'id' and 'text'.\\ Analyze each text and provide the required classification.
\end{tcolorbox}
\label{prompt:multilabel}    
\end{figure}
\begin{figure}[h!]
    \centering
    \begin{tcolorbox}[colback=gray!5!white, colframe=black, title= In-context Examples in Few-Shot Prompting for Binary Classification, width=0.47\textwidth]
\small
\textbf{HATE(true)}\\
\noindent 1. \emph{Ei hindu gula ke desh theke ber kore dite hobe, era shobi desher shotru}\\
\noindent \textbf{English Translation }: These Hindus need to be thrown out of the country, they are all enemies of the nation.\\
\noindent \textbf{Reason}: Calls for the expulsion of religious groups (Hindus) and labels them all as enemies\\

\noindent 2. \emph{Chakma der manush e bhabo na, oder
shob ke mere fela uchit}\\
\noindent \textbf{English Translation :} Don’t even think of the Chakmas as humans, all of them should be killed.\\
\noindent \textbf{Reason}: Dehumanizes ethnic group (Chakma) and calls for violence against them\\

\textbf{NON-HATE(False)}\\
\noindent 1. \emph{Ami or political opinion er shathe ekmat hoi na, kintu sheta niye kharap lage na}\\
\noindent \textbf{English Translation:} I don’t agree with his political opinion, but it doesn’t bother me.\\
\noindent \textbf{Reason}: Expresses disagreement without hatred or discrimination\\

\noindent 2. \emph{Amader team abar harlo, ki kharap khelse ora}\\
\noindent \textbf{English Translation}: Our team lost again, they played so badly\\
\noindent \textbf{Reason}: Expresses criticism without targeting any protected group
\label{prompt:few_shot_binary_examples}

\end{tcolorbox}
\end{figure}

\begin{figure}[h!]
    \centering
\begin{tcolorbox}[colback=gray!5!white, colframe=black, title= CoT-Based Prompt, width=0.47\textwidth]
\small
CHAIN-OF-THOUGHT ANALYSIS STEPS:\\
1. Initial reading: Read the entire text and identify any potential hate speech elements.\\
2. Keyword identification: Highlight specific words or phrases that could indicate hate speech.\\
3. Context analysis: Consider the overall context and tone of the message.\\
4. Category mapping: Connect the identified elements to specific hate speech categories.\\
5. Multi-label consideration: Determine if the text fits into multiple categories.\\
6. Intensity assessment: Evaluate the severity or intensity of the hate speech, if present.\\
7. Final classification: Summarize the findings and assign relevant categories.\\

\end{tcolorbox}
\label{prompt:CoT}    
\end{figure}
\begin{figure}[h!]
    \centering

\begin{tcolorbox}[colback=gray!5!white, colframe=black, title= In-context Examples in Few-Shot Prompting for Multi-label Classification, width=0.47\textwidth]
\small
\textbf{HATE(true)}\\
\noindent 1. \emph{Ei hindu neta gula ke desh theke ber kore dite hobe, era shobi desher shotru}\\
\noindent \textbf{English Translation}: These Hindu leaders need to be thrown out of the country, they are all enemies of the nation.\\
\noindent \textbf{Categories}: RELIGIOUS, POLITICAL\\
\noindent \textbf{Reason}: Calls for expulsion of religious group (Hindus) and labels them all as enemies\\

\noindent 2. \emph{Tumi ekta mota beyadob, tomar moto lok der vote deoya uchit na}\\
\noindent \textbf{English Translation}:You are a rude fat person, people like you don’t deserve to be voted for.\\
\noindent \textbf{Categories}: BODY SHAMING, POLITICAL, PERSONAL OFFENCE\\
\noindent \textbf{Reason}: Uses body shaming language, suggests denying voting rights, and personally insults the individual\\

\noindent 3. \emph{Ei mohila neta ra shudhu ghor e thakle e desh er unnoti hobe, oder politics e ashar ki dorkar}\\
\noindent \textbf{English Translation}: If these women leaders stayed at home, the country would progress; why do they need to come into politics?\\
\noindent \textbf{Categories}: GENDER, POLITICAL\\
\noindent  \textbf{Reason}: Discriminates against women in politics, targeting both gender and political participation\\

\textbf{NON-HATE(False)}\\
\noindent 1. \emph{Ami or political opinion er shathe ekmat hoi na, kintu sheta niye kharap lage na}\\
\noindent \textbf{English Translation:} I don’t agree with his political opinion, but it doesn’t bother me.\\
\noindent \textbf{Categories}: None\\
\noindent \textbf{Reason}: Expresses disagreement without hatred or discrimination\\

\noindent 2. \emph{Amader team abar harlo, ki kharap khelse ora}\\
\noindent \textbf{English Translation}: Our team lost again, they played so badly\\
\noindent \textbf{Categories}: None\\
\noindent \textbf{Reason}: Expresses criticism without targeting any protected group
\label{prompt:few_shot_multilabel_examples}
\end{tcolorbox}
\end{figure}
\begin{figure}[h!]
    \centering
\begin{tcolorbox}[colback=gray!5!white, colframe=black, title= Translation-Based Prompt, width=0.47\textwidth]
\small
You are an expert at detecting hate speech. You will be given an array of Bangla texts (written in Latin script/English letters). First, convert the Bangla Transliterated text into standard Bangla without losing the semantic meaning, then classify the text as either hate speech (true) or non-hate speech (false). Provide a reason for your classification and provide the translation.

\end{tcolorbox}
\label{prompt:TranslationBased}   
\end{figure}

\clearpage


\end{document}